\documentclass[12pt, a4paper, twoside, openright]{book} 
\usepackage[top=3cm, bottom=3cm, left=3cm, right=3cm]{geometry}

\usepackage[english]{babel} 
\usepackage[T1]{fontenc}
\usepackage[utf8x]{inputenc} 
\usepackage{mathtools} 
\usepackage{graphicx} 
\usepackage{url} 
\usepackage{booktabs,subcaption} 
\usepackage{amsthm,amsmath,amssymb,mathrsfs,dsfont}
\usepackage{rotating}
\usepackage[font=itshape,noorphans]{quoting}
\usepackage{bm,bbm}
\usepackage{microtype}
\microtypesetup{protrusion=true,expansion=true}

\usepackage{hyperref}
\usepackage[dvipsnames]{xcolor}

\usepackage{tikz}
\usetikzlibrary{shapes.geometric,arrows.meta,positioning,fit}
\tikzset{
  box/.style = {draw, rounded corners, minimum width=28mm, minimum height=7mm, align=center, font=\small},
  subbox/.style = {draw, rectangle, minimum width=24mm, minimum height=6mm, align=center, font=\footnotesize},
  arr/.style = {-{Latex[length=3mm]}, very thick}
}

\usepackage{scrextend}
\usepackage[ruled,norelsize,vlined]{algorithm2e}

\usepackage[round]{natbib}
\usepackage[nottoc]{tocbibind}

\usepackage[small]{titlesec}

\usepackage{caption}
\usepackage[round]{natbib}
\usepackage{setspace}
\usepackage[all]{nowidow} 
\finalhyphendemerits=\doublehyphendemerits

\usepackage[osf]{mathpazo} 
\usepackage[euler-digits]{eulervm}
\usepackage[scaled=0.85]{beramono}
\DeclareSymbolFont{greekletters}{OML}{cmr}{m}{it}

\definecolor{webgreen}{rgb}{0,0.5,0}
\definecolor{webbrown}{rgb}{0.6,0,0}
\definecolor{structural}{RGB}{0,0, 130}   

\usepackage{fancyhdr}

\newenvironment{acknowledgements}%
{\cleardoublepage\null\vfill\begin{center} %
\bfseries Ringraziamenti\end{center}} %
{\vfill\null}

\newtheoremstyle{classicdef}
{12pt}
{12pt}
{}
{}
{\scshape}
{:}
{.5em}
{}

\theoremstyle{definition}

\numberwithin{assumption}{chapter}

\theoremstyle{definition}

\numberwithin{definition}{chapter}

\theoremstyle{remark}

\numberwithin{example}{chapter}

\theoremstyle{remark}

\numberwithin{remark}{chapter}

\newtheoremstyle{classicthm}
{12pt}
{12pt}
{\itshape}
{}
{\scshape}
{:}
{.5em}
{}

\theoremstyle{plain}

\numberwithin{theorem}{chapter}

\numberwithin{corollary}{chapter}

\numberwithin{lemma}{chapter}

\numberwithin{proposition}{chapter}

\hypersetup{%
    colorlinks=true, linktocpage=true, pdfstartpage=1, 
    breaklinks=true, pdfpagemode=UseNone, pageanchor=true, pdfpagemode=UseOutlines,%
    plainpages=false, bookmarksnumbered, bookmarksopen=true, bookmarksopenlevel=1,%
    hypertexnames=true, pdfhighlight=/O,
    urlcolor=webbrown, linkcolor=Maroon, citecolor=structural, 
}

\graphicspath{{img/}} 

\begin{document}

\frontmatter

\pagestyle{empty}

\pagestyle{empty} 

\begin{titlepage}
  
 \begin{center}
 {\large  

 \hfill

 \vfill
 {
 {\Large \textsc{Università degli studi di Milano--Bicocca}}\\
 {\textsc{Scuola di Economia e Statistica}}\\
 \vfill
	\textsc{Corso di Laurea  in} \\
	\textsc{Scienze Statistiche ed Economiche} \\
	\vfill
	\includegraphics[width=4cm]{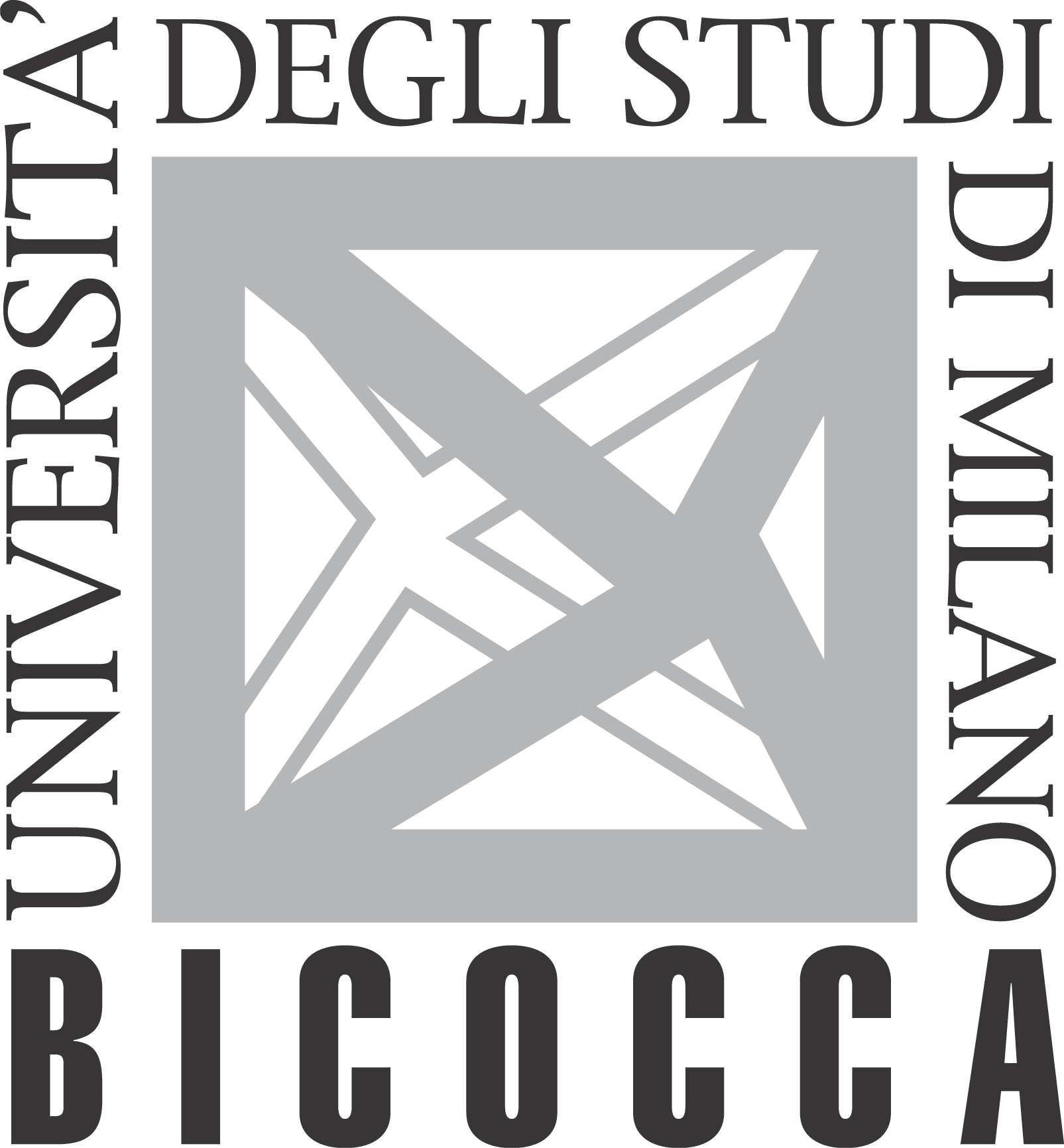}
 \vfill
 
 {\Huge\color{Maroon}\textsc{Transformers for Tabular Data: A Training Perspective of Self-Attention via Optimal Transport}}\\
 }
}
\end{center}

\vfill
{
\large
\begin{flushleft}
\textsc{Relatore}: Prof. Antonio Candelieri \\
\textsc{Correlatore}: Prof. Matteo Maria Pelagatti \\
\end{flushleft}

\vfill
\begin{flushright}
\textsc{Tesi di laurea di}:\\
 Alessandro Quadrio\\
\textsc{Matricola N. 871554}
\end{flushright}

\vfill
\begin{center}
\textsc{Anno Accademico 2024/2025}
\end{center}

}
\end{titlepage}

\pagestyle{empty}

\cleardoublepage
\begin{center}
    {\Large \bfseries Abstract}
\end{center}
\vspace{1.5em}
\begin{center}
\setlength{\parindent}{0pt}
\setlength{\parskip}{0.9em}

This thesis examines self-attention training through the lens of Optimal Transport (OT) and develops an OT-based alternative for tabular classification.
The study tracks intermediate projections of the self-attention layer during training and evaluates their evolution using discrete OT metrics, including Wasserstein distance, Monge gap, optimality, and efficiency.
Experiments are conducted on classification tasks with two and three classes, as well as on a biomedical dataset.

Results indicate that the final self-attention mapping often approximates the OT optimal coupling, yet the training trajectory remains inefficient. Pretraining the MLP section on synthetic data partially improves convergence but is sensitive to their initialization.
To address these limitations, an OT-based algorithm is introduced: it generates class-specific dummy Gaussian distributions, computes an OT alignment with the data, and trains an MLP to generalize this mapping.
The method achieves accuracy comparable to Transformers while reducing computational cost and scaling more efficiently under standardized inputs, though its performance depends on careful dummy-geometry design.
All experiments and implementations are conducted in R.
\end{center}


\tableofcontents

\begin{acknowledgements}

Gli eventuali meriti di questa tesi si devono alla pazienza e alla fiducia del mio relatore Antonio Candelieri;

\vspace{0.5em}

ai miei genitori Gea e Luca e a mia sorella Sara per cui il duro lavoro è il principio di ogni proposito;

\vspace{0.5em}

al lungo tempo trascorso con Alessandro, Carolina, Daniel, Domenico, Lorenzo, Nicolò, Riccardo e Vanessa il cui affetto sincero mi commuove;

\vspace{0.5em}

e soprattutto ad Alice che rende ogni attimo indimenticabile e ogni ora un sogno.

\end{acknowledgements}

\vspace{1cm}
\begin{center}
\textit{in questo mondo la bellezza è comune} \\
--- Jorge Luis Borges
\end{center}
\vspace{3cm}

\cleardoublepage

\mainmatter

\pagestyle{fancy}

\chapter{Introduction}

Since their introduction in \textit{Attention is All You Need} \cite{vaswani2017attention}, Transformers have become the dominant architecture across a wide range of tasks, particularly in Natural Language Processing, and increasingly in computer vision \cite{drones7050287}, speech processing \cite{transformersspeechprocessingsurvey}, and multi-modal applications \cite{multimodal}. Their success has stimulated extensive research, both in adapting the architecture to new domains and in probing the mechanisms underlying their remarkable performance \cite{badaro2023transformers}.

A key factor behind the Transformer’s effectiveness is its high degree of overparameterization. While this makes training computationally demanding, it allows the model to approximate highly complex functions, often achieving state-of-the-art results. This has driven the development of ever-larger models, frequently exceeding hundreds of billions of parameters, with training costs rising correspondingly.

Despite their empirical success, the theoretical understanding of Transformers remains limited. Fundamental questions, such as the dynamics of training at scale and the properties of the Self-Attention (SA) mechanism, are still not fully resolved. Additionally, computational efficiency continues to be a major challenge, raising the question of whether conventional stochastic gradient-based training is the most effective approach.

Consequently, recent research has pursued alternatives to straightforward model scaling, either by improving efficiency through techniques such as pruning \cite{pruning}, distillation \cite{distilleria}, and compression \cite{compression}, or by revisiting the underlying principles of the architecture. In this work, the latter perspective is adopted, analyzing the Transformer through the lens of Optimal Transport (OT) theory (\cite{peyre2019computational}, \cite{peyre2025optimal}, \cite{OTambrogio}).

The SA mechanism can be interpreted as transporting probability mass between input representations (tokens in NLP, or features in other settings) and their transformed outputs. OT provides a rigorous mathematical framework to formalize this notion, offering a novel perspective on how attention redistributes information and suggesting potential avenues for algorithmic improvement.

This work makes two main contributions. First, I analyze SA using OT to study how it progressively remaps input data during training. To this end, the model is trained using full-batch (non-stochastic) backpropagation, and the intermediate projections generated by the SA layer at each epoch are recorded. These projections form a trajectory in latent space, capturing the evolution of representations throughout training. Applying OT to this trajectory allows for a detailed characterization of how SA organizes information and adapts the latent structure of the inputs.

Second, I propose an alternative Transformer-inspired algorithm that directly leverages OT principles. This method is designed to improve training efficiency and reduce computational cost while retaining the core functional capabilities of SA. Although this thesis focuses on tabular data, the approach is general and can be extended to other modalities with minimal modification.

By combining the theoretical framework of Optimal Transport with empirical analysis, this thesis aims to deepen our understanding of the internal dynamics of Transformers and explore more efficient training strategies grounded in principled mathematical foundations.

The R implementation of all models and simulation setups is available in an open-access GitHub repository \footnote{\url{https://github.com/AleQuad43/self-attention-optimal-transport}}.

\chapter{Methodological Background}

\section{Transformers for Tabular Data}
Over the past decades, a variety of deep learning architectures have been developed for representation learning, from convolutional networks to GANs and VAEs, and more recently Transformers—which have since become the dominant architecture across a wide range of tasks \cite{aggarwal2018neural,vaswani2017attention}—and which were originally designed for sequence-to-sequence tasks in Natural Language Processing (NLP).
In this context, the model’s purpose is to map a sequence of tokens in one language to a sequence in another.
Since the Transformer actually operates on embeddings: points in a high-dimensional space, the application of Transformers to tabular data is natural.
Whether the inputs are words, pixels, or tabular features, the architecture only requires that each input element be embedded into a continuous representation. 
There is no explicit dependence on sequential or spatial structure unless, as it is common practice in NLP, one uses positional encodings: an extra feature that tracks the positional relationships between tokens in the text.

\subsection*{The Encoder Architecture}

\begin{figure}
\centering
\includegraphics[width=0.5\linewidth]{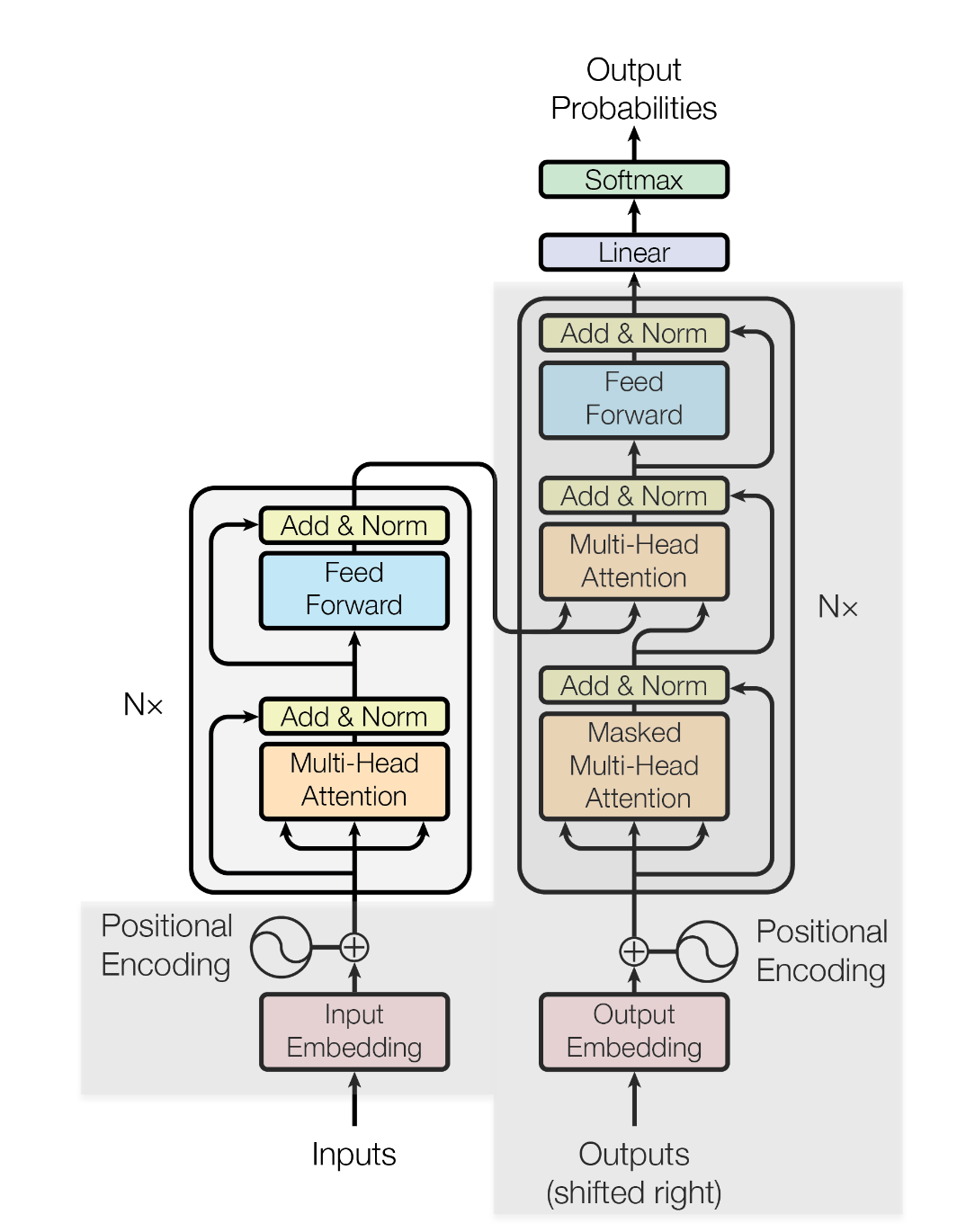}
\caption{Transformer architecture (adapted from \cite{vaswani2017attention}). 
The decoder and other components not relevant for tabular data are grayed out, 
as this work focuses exclusively on the encoder.}
\label{fig:attention is gay}
\end{figure}

In the original formulation, the Transformer is composed of an encoder and a decoder ( Figure \ref{fig:attention is gay}). The encoder maps input embeddings into contextualized representations, and the decoder generates an output sequence based on these representations. In NLP, this structure is crucial, since the task requires transforming an input sentence into an output sentence. Outside of language tasks, however, the decoder loses much of its purpose: for classification or regression on tabular data, one only needs to transform the input into a useful representation for prediction. For this reason, in this thesis, only the encoder component is employed.

Each encoder block consists of two main components: a SA mechanism and a Multilayer Perceptron network (MLP), both wrapped in residual connections and layer normalization. The residual connections play a critical role ensuring that information can propagate across layers without degradation, and mitigating the vanishing gradient problem. Empirically, they allow the network to retain direct access to the input representation while simultaneously learning complex transformations.

Having introduced the overall encoder architecture, I now turn to the mathematical formulation of SA, which will be central to both the forthcoming analysis and the alternative model proposed in this thesis.

\subsection*{Self-Attention}
\label{ref:Self_Ateention}

The central innovation of the Transformer is the SA mechanism, which enables the model to learn dependencies between input elements. Given an input matrix $X \in \mathbb{R}^{t \times p}$, where $t$ is the number of input elements (tokenised words in a sentence in NLP) and $p$ is the number of features, three linear projections are computed:
\[
    Q = XW_Q, \quad K = XW_K, \quad V = XW_V,
\]
with learnable parameter matrices $W_Q, W_K, W_V \in \mathbb{R}^{p \times d}$.  
The matrices $Q, K, V \in \mathbb{R}^{t \times d}$ are called Query, Key, and Value. The SA output is then given by
\begin{equation}
\label{eqaution:self_attention}
    \text{Attention}(Q,K,V) = \text{softmax}\!\left(\frac{QK^\top}{\sqrt{d}}\right) V
\end{equation}

This formulation can be interpreted as computing similarity scores through queries and keys, normalizing them into a probability distribution, and using this distribution to form weighted combinations of the values. As will be discussed in detail in Section~\ref{sec:ot_analysis}, SA can be understood as a remapping of the input data within the same latent space. This perspective highlights how attention facilitates the feed-forward subsequent layer by reshaping the representation in a way that makes the predictive task more tractable.

In practice, multiple attention heads are used in parallel (multi-head attention), each with independent projection matrices. This allows the model to attend to different aspects of the data simultaneously, enabling it to learn more complex transformations. Formally, given $h$ heads, the output of the $i$-th head is
\[
    \text{head}_i = \text{Attention}(QW_Q^{(i)},\, KW_K^{(i)},\, VW_V^{(i)}),
\]
where $W_Q^{(i)}, W_K^{(i)}, W_V^{(i)} \in \mathbb{R}^{p \times d}$ are learnable projection matrices for the $i$-th head and     $\text{head}_i \in \mathbb{R}^{t \times p}$.

The outputs of all heads are then concatenated and projected back into the model dimension $d$ through a learned weight matrix $W^O \in \mathbb{R}^{hd \times p}$:
\[
    \text{MultiHead}(Q,K,V) = \text{Concat}\big(\text{head}_1, \ldots, \text{head}_h\big) W^O.
\]

The resulting representation is subsequently passed through a feed-forward sublayer (FF), which remains part of the SA layer and is structurally distinct from the MLP mentioned before.  
The MLP, in contrast, is the block responsible for producing task-specific predictions and is applied after the SA layers. 

While FF and MLP share the same underlying structure, the key differences are that FF includes a residual connection and operates within the SA layer, whereas the MLP performs prediction using the SA output as input, with its architecture depending on the task (e.g., ending with a softmax layer for classification, as employed in the experiments of this thesis).

\section{Optimal Transport}

Optimal Transport (OT) is a mathematical framework for comparing and transforming probability distributions. At its core, OT seeks the most cost-efficient way to move mass from one probability distribution to another. This concept has its origins in a problem first formulated by Gaspard Monge in the 18th century, where the goal was to minimize the total cost of transporting a pile of soil to fill a given excavation. In modern terms, given two probability measures $\mu$ and $\nu$ defined on spaces $\mathcal{X}$ and $\mathcal{Y}$, respectively, OT formalizes the idea of transporting mass from $\mu$ to $\nu$ under a chosen cost function $c(x,y)$.

\subsection*{Monge problem (arbitrary measures)}

There are two principal formulations of the OT problem. The first one is the \emph{Monge problem} which seeks a deterministic map $T$ that  transports a measure ($\alpha=\sum_{i=1}^n a_i \delta_{x_i}$) to another measure ($\beta=\sum_{i=1}^n b_i \delta_{x_i}$) under the constraint of \emph{mass preservation}
Formally we want to solve:
\[ 
\min_{T}\;\left\{ \sum_{i=1}^n c\bigl(x_i,\,T(x_i)\bigr)\;:\; T_{\#}\alpha = \beta \right\}.
\]

The constraint $T_{\#}\alpha=\beta$ (\emph{mass preservation}) imposes that $T$ must redistribute the mass of $\alpha$ exactly to match $\beta$. 
This formulation can sometimes be restrictive: the map $T$ may not exist in general, especially when $\alpha$ and $\beta$ are not compatible in terms of support or weight allocation.

\paragraph{Discrete uniform case.}  
In our application, we are concerned with point clouds of equal cardinality, therefore the formulation of the problem simplifies to

\[
    \min_{\sigma \in \text{Perm}(n)} \frac{1}{n} \sum_{i=1}^n c(x_i, y_{\sigma(i)})
\]
in the case $\alpha = \frac{1}{n}\sum_{i=1}^n \delta_{x_i}$ and $\beta = \frac{1}{n}\sum_{j=1}^n \delta_{y_j}$.

Since both measures are uniform, the mass of each point is identical and equal to $1/n$. The Monge problem then reduces to finding a bijection $\sigma \in \text{Perm}(n)$ that pairs each $x_i$ with exactly one $y_{\sigma(i)}$:

This formulation is also known as the \emph{optimal assignment problem} and corresponds to solving a minimum-cost matching between two point clouds.

\subsection*{Kantorovich relaxation (discrete measures)}

The Monge problem imposes a strong restriction: each mass element must be transported in its entirety to a unique target. This deterministic formulation is not always feasible. To address this, Kantorovich proposed a relaxation where transport can be probabilistic, allowing mass splitting.

\begin{figure}[h]
    \centering
    \includegraphics[width=1\linewidth]{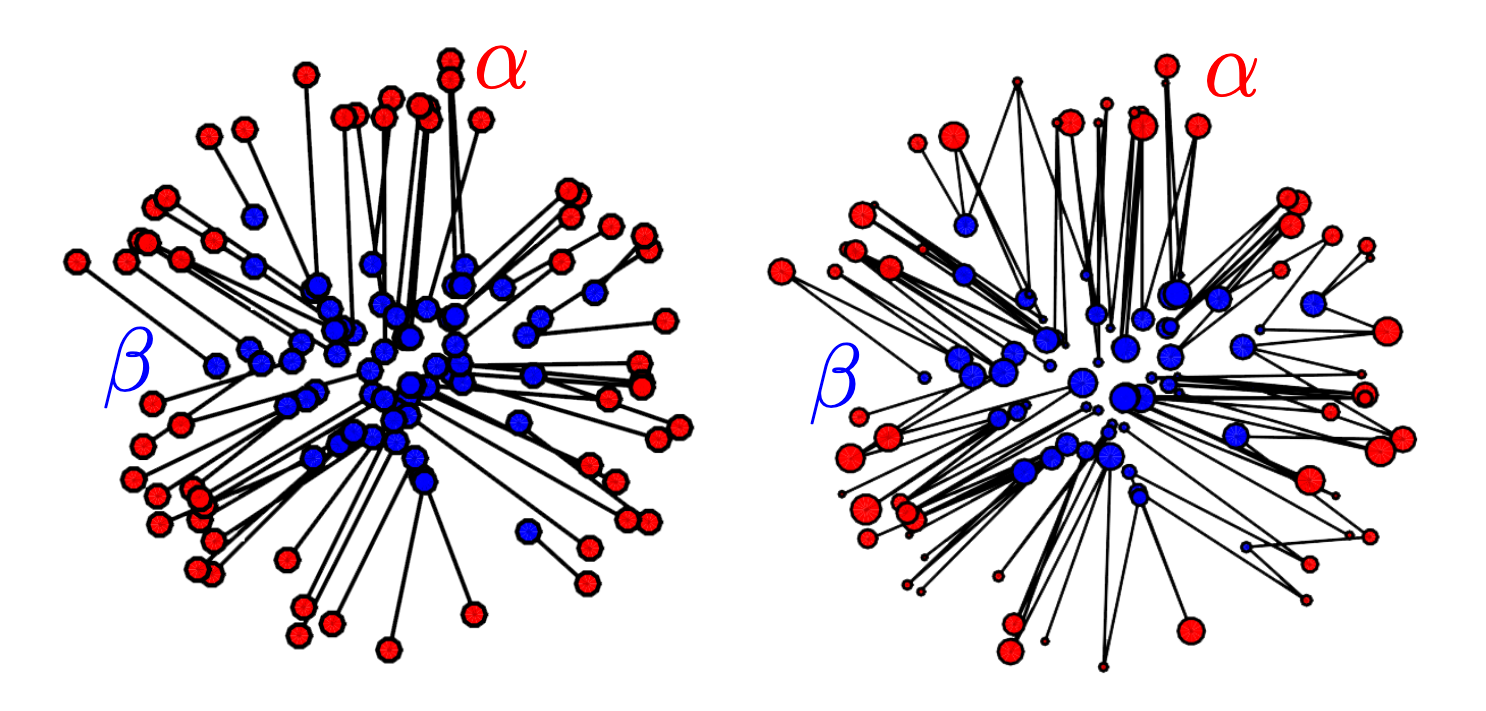}
\caption{Illustration of the difference between the Monge formulation (left) and the Kantorovich relaxation (right). In the Monge case, when both measures have the same number of points and equal weights, the optimal plan corresponds to a permutation matrix that pairs each $x_i$ with exactly one $y_j$. In contrast, the Kantorovich formulation allows for probabilistic couplings between arbitrary discrete measures that cannot be matched one-to-one. Black segments indicate the nonzero entries of the optimal coupling matrix $P$. Figure adapted from \textit{Computational Optimal Transport} by Gabriel Peyré and Marco Cuturi.}
\label{fig:monge-kantorovich}
    \label{fig:placeholder}
\end{figure}

For discrete measures of the form
\[
    \alpha = \sum_{i=1}^n a_i \, \delta_{x_i}, 
    \qquad 
    \beta = \sum_{j=1}^m b_j \, \delta_{y_j},
\]
with $a \in \Sigma_n$, $b \in \Sigma_m$ (the simplices of dimension $n$ and $m$), an admissible transport plan is represented by a coupling matrix $P \in \mathbb{R}_+^{n \times m}$ such that
\[
    P \mathbf{1}_m = a,
    \qquad 
    P^\top \mathbf{1}_n = b.
\]
The Kantorovich optimal transport problem then reads
\[
    L_C(a,b) = \min_{P \in U(a,b)} \langle C, P \rangle 
    = \min_{P \in U(a,b)} \sum_{i=1}^n \sum_{j=1}^m C_{i,j} P_{i,j},
\]
where $C_{i,j} = c(x_i,y_j)$. Unlike the Monge problem, this is a convex linear program, always admitting a solution.
Figure \ref{fig:monge-kantorovich} shows a graphical representation of the difference between the \emph{Monge problem} and the \emph{Kantorovich relaxation}.

\clearpage

\subsection*{Wasserstein distances (between discrete measures)}

A central notion in OT is the \emph{Wasserstein distance}, which lifts a ground metric between points to a metric between probability measures. For $p \geq 1$ and cost $c(x,y)=d(x,y)^p$, the $p$-Wasserstein distance between measures $\alpha$ and $\beta$ is defined as
\[
    W_p(\alpha, \beta) 
    = \    L_C(\alpha,\beta) = \min_{P \in U(\alpha,\beta)} \langle C, P \rangle
\]
Here $U(\alpha,\beta)$ denotes the set of admissible couplings (joint distributions with marginals $\alpha$ and $\beta$).

If $D \in \mathbb{R}^{n \times n}_+$ is a distance on $(1,\dots, n)$, which is to say that it satisfies the following properties:
\begin{itemize}
    \item[(i)] $D \in \mathbb{R}^{n \times n}_+$ is symmetric;
    \item[(ii)] $D_{i,j} = 0$ if and only if $i=j$;
    \item[(iii)] $\forall (i,j,k) \in (1,\dots, n)^3, \quad D_{i,k} \leq D_{i,j} + D_{j,k}$,
\end{itemize}
then for $p \geq 1$ and $C = D^p$, the $p$-Wasserstein distance
\[
    W_p(\alpha,\beta) \ \overset{\text{def.}}{=} \ L_{D^p}(\alpha,\beta)^{1/p}
\]
inherits the same properties and is itself a distance on the space of probability measures $\Sigma_n$.

Since, under the \emph{Kantorovich relaxation}, the optimal transport problem can be solved between any two probability measures, the \emph{Wasserstein distance} provides a well-defined metric on the space of probability measures, extending the notion of distance from individual points to entire distributions. In particular, the \emph{Wasserstein distance} is naturally defined not only between two discrete measures, but also between a discrete and a continuous distribution (unlike, for example, the \emph{Kullback--Leibler divergence}). This property has led to the widespread adoption of the \emph{Wasserstein distance} in various fields of Machine Learning (most famously the Wasserstein GAN \cite{arjovsky2017wasserstein}). 

One limitation of the \emph{Wasserstein distance} is its reliance on the choice of ground cost. If the cost function does not provide a meaningful notion of dissimilarity between data points, then the Wasserstein distance itself carries little information. This issue arises, for instance, when using the Euclidean distance on a dataset that combines continuous and categorical variables: the Euclidean distance on dummy variables is not directly comparable with the same distance on continuous features.

\chapter{Analysis of Self-Attention with Optimal Transport}
\label{sec:ot_analysis}

As shown in \ref{ref:Self_Ateention}, SA is the central component of the Transformer architecture; through over-parameterization and residual connections, it enables the model to approximate highly complex functions of the input data.

Formally, the operation in Equation~\eqref{eqaution:self_attention} produces a new representation of the input by applying a stochastic matrix 
\[
    P = \text{softmax}\!\left(\tfrac{QK^\top}{\sqrt{d}}\right) \in \mathbb{R}^{t \times t}
\]
to the value matrix $V$. Each row of $P$ defines a probability distribution over the input elements (same features at different time steps), specifying how the learned representation of a given point is expressed as a convex combination of input data.
In this sense, the mechanism does not merely extract features but \emph{remaps} the input data according to learned similarity relations.  

Interestingly, while $P$ is by construction row-stochastic, recent work has shown that during training it tends to converge towards a \emph{doubly stochastic} matrix \citep{sander2022sinkformers}.

This remapping has the effect of concentrating relevant information and filtering out less useful variation, so that the subsequent MLP layers are exposed to inputs that are already organized in a task-relevant manner. Consequently, the MLP layers operate on representations where dependencies have been made explicit, reducing the complexity of the mapping they need to learn.

\begin{figure}[h]
  \centering
  \begin{minipage}{0.48\linewidth}
    \centering
    \includegraphics[width=\linewidth]{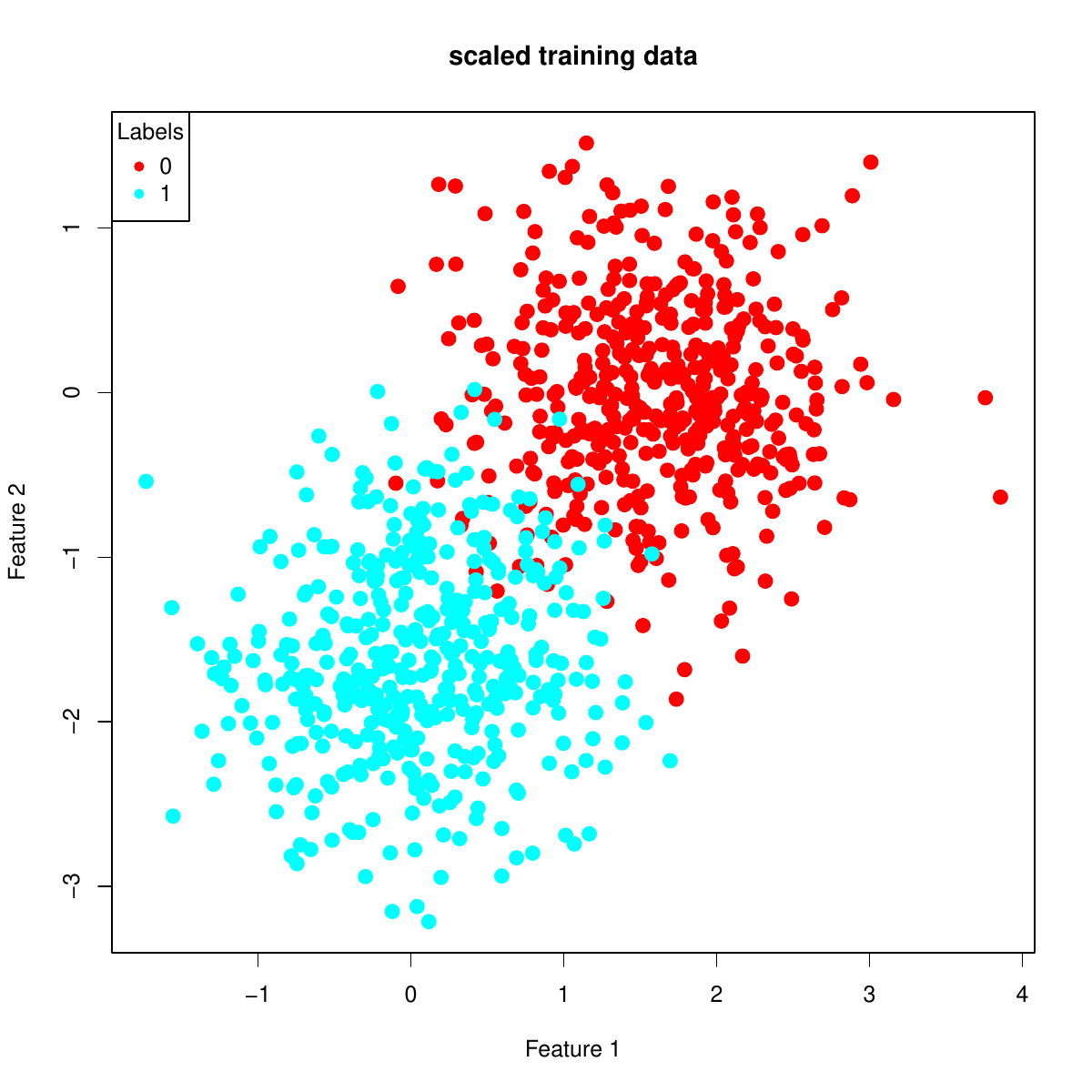}
  \end{minipage}
  \hfill
  \begin{minipage}{0.48\linewidth}
    \centering
    \includegraphics[width=\linewidth]{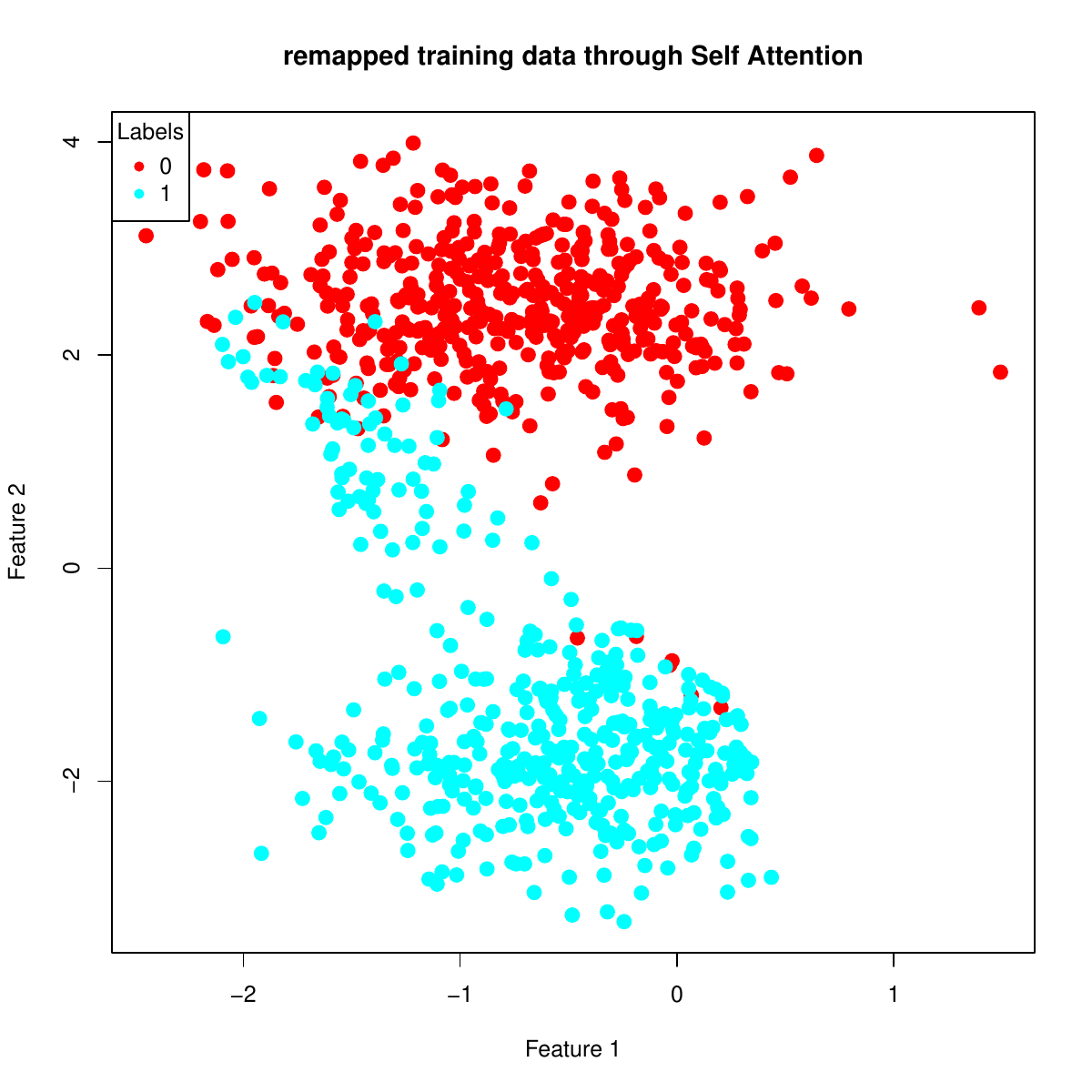}
  \end{minipage}

  \caption{Example of learned remapping of the SA after 45 iterations. On the left the input data (scaled), on the right the learned remapping.}
  \label{fig:transformer-remap}
\end{figure}

Given this, training the SA block can be understood as searching for the optimal remapping of the inputs to perform prediction.  
In the case of classification, this amounts to finding a remapping that arranges the latent representations of the input data so that the MLP can easily separate them across classes (Figure \ref{fig:transformer-remap}).

Today this remapping is learned with stochastic gradient descent (SGD), as is common in deep learning.
Viewed step by step, each iteration of SGD modifies the SA parameters so that the induced stochastic matrix $P$ gradually reshapes the latent representation of the inputs.
In other words, the training process can be seen as progressively transporting probability mass in the latent space to achieve an effective remapping of the data.  

This perspective naturally connects to OT: by examining the evolution of latent representations through OT, we can quantify how close this path is to an optimal one.

A possible workaround is offered by Wasserstein Gradient Flow (\cite{santambrogio2017euclidean}), but a detailed exploration of this direction is left for future work; see \cite{lambert2022variational}, which develops variational inference methods grounded in Wasserstein gradient flows, and \cite{global}, which analyzes convergence properties of gradient flows in Transformers.

see \cite{global}, which analyzes the convergence properties of gradient flow in Transformers through this framework.

Instead, I will take the SA solution as given (justified by the record-breaking performance of Transformers) and focus on understanding whether it was reached in an efficient and effective way.

\section{Experiments}

To investigate the OT properties of SA, the experiments were designed as 2- and 3-label classification tasks.
The methodology was as follows.  
The data of each label were generated from independent Gaussians, with samples being i.i.d., and then arranged in a dataset 
$\mathbf{X} \in \mathbb{R}^{n\times t \times p}$, 
where $n$ is the number of instances, $t$ the number of sampled batches and $p$ the number of features (fixed at 2).  
The difficulty of the classification task was measured by the Wasserstein distance between the Gaussians from which the data were generated.

I then trained the Transformers (training parameters are displayed together with the results) with a small modification in order to extract the SA remapping at each iteration of gradient descent. Note that the training algorithm was not stochastic, since the entire training dataset was used at each iteration. The data were standardized before training.  

\begin{figure}[h]
  \centering
  \begin{minipage}{0.48\linewidth}
    \centering
    \includegraphics[width=\linewidth]{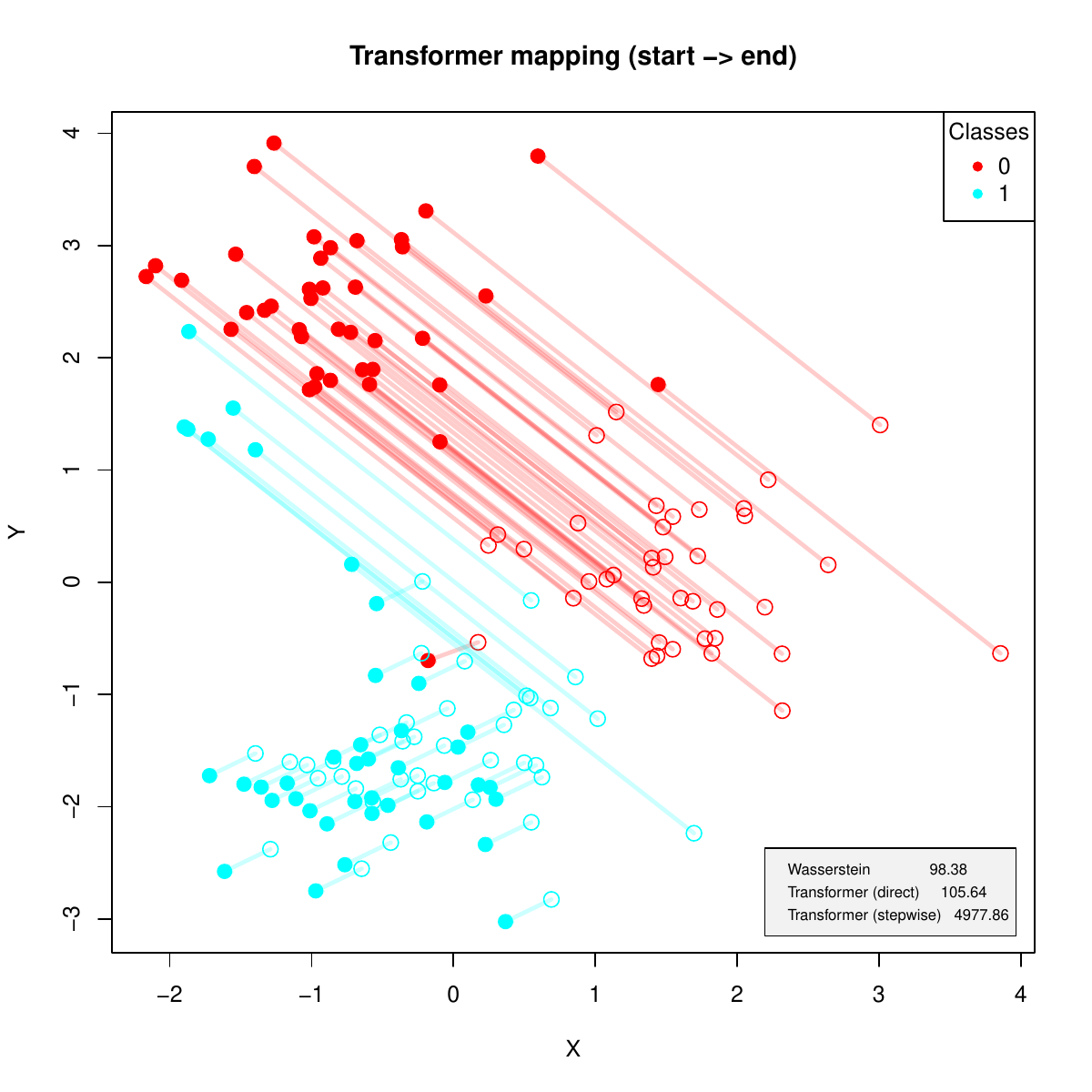}
  \end{minipage}
  \hfill
  \begin{minipage}{0.48\linewidth}
    \centering
    \includegraphics[width=\linewidth]{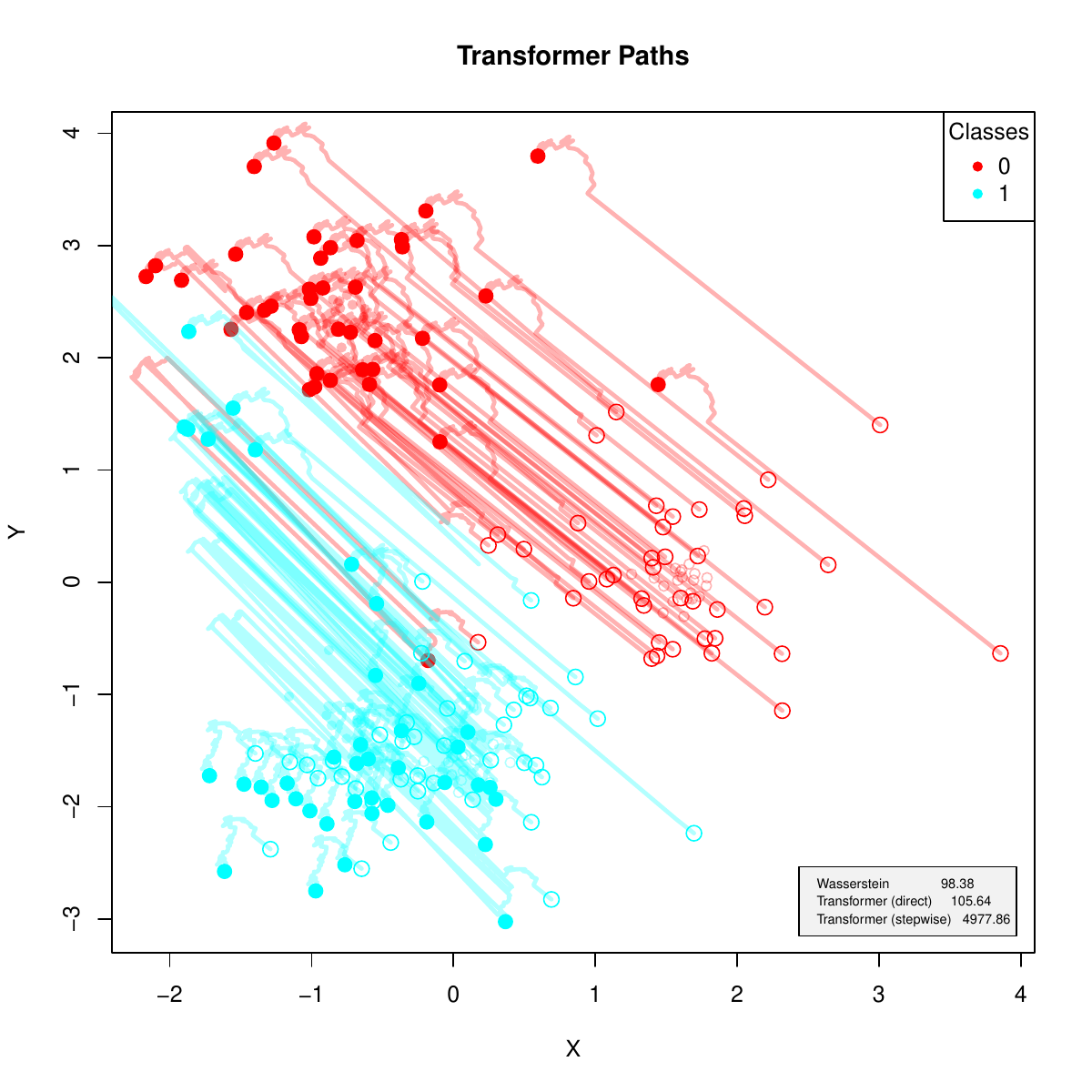}
  \end{minipage}
\caption{Transformer-induced transport: on the left, the SA matching between input points (empty) and their final remapped positions (full); on the right, the full Transformer path showing the sequence of projections through training.}
  \label{fig:transformer OT solved}
\end{figure}

After training, the projections after the best epoch (i.e., the epoch that achieved the minimum training loss) were not retained.
I then solved, class by class, a discrete optimal transport problem between the original (scaled) input data and their final remapped positions (best-epoch projection), using the short simplex algorithm.
This was done class by class as shown in Figure \ref{fig:transformer OT solved}.
At last the following metrics were computed:

\begin{itemize}
    \item \textbf{Accuracy (point-wise)}: fraction of correctly classified points (each time step in each instance);
    \[
        \text{Acc}_{\text{point}} 
        = \frac{1}{nt} \sum_{i=1}^n \sum_{s=1}^t 
        \mathbb{1}\!\left\{ y_{i,s} = \arg \max \hat{p}_{i,s}(c) \right\},
    \]
    where
    \begin{itemize}
        \item $n$ is the number of instances,
        \item $t$ is the number of time steps per instance,
        \item $C$ is the number of classes,
        \item $y_{i,s} \in \{1, \dots, C\}$ is the true class of point $(i,s)$,
        \item $\hat{p}_{i,s}(c)$ is the predicted probability (softmax output) that point $(i,s)$ belongs to class $c$,
        \item $\mathbb{1}\{\cdot\}$ is the indicator function.
    \end{itemize}

    \item \textbf{Accuracy (instance-wise)}: fraction of correctly classified entire instances (all time steps considered together);
    \[
        \text{Acc}_{\text{inst}} 
        = \frac{1}{n} \sum_{i=1}^n
        \mathbb{1}\!\left\{ y_{i} = \arg \max \frac{1}{t}\sum_{s=1}^t \hat{p}_{i,s}(c) \right\},
    \]
    where $y_i$ is the true class of instance $i$.
    
    \item \textbf{Computational time} for the entire training procedure;
    \item \textbf{Matching}: the proportion of points whose Transformer-induced matching coincided with the OT matching (a value close to 1 means that most points were transported optimally, while 0 means none were);
    \item \textbf{Wasserstein distance} between the two point clouds (input vs.\ remapped representations);
    \item \textbf{Transformer distance}: the straight-line distance between input points and their corresponding Transformer-induced matches;
    \item \textbf{Transformer cost}: the cumulative transport cost along the full sequence of projections, i.e.\ through every remapping at each training step.
    \item \textbf{Monge Gap}: the difference between the Transformer distance and the Wasserstein distance, quantifying deviation from the OT-optimal transport, this measure was introduced in \citep{monge_gap}; 
    \item \textbf{Optimality}: the ratio of Wasserstein distance over Transformer distance, providing an alternative measure of how close the transport was to optimal (similar interpretation to the Monge Gap);
    \item \textbf{Efficiency}: the ratio of Transformer distance over Transformer path; a value close to 1 indicates a nearly straight path to the solution, while close to 0 indicates a highly wandering trajectory;

    \item \textbf{Best epoch}: the epoch achieving minimum training loss;
    \item \textbf{Recall of each class}: standard recall computed separately for each label computed instance-wise.
\end{itemize}

Each simulation was conducted with 100 repetitions.

All experiments were run on a MacBook Pro under macOS Big Sur (Intel x86\_64, R 4.2.2), using CPU only (no GPU acceleration).

\subsection{Regular Transformer}

All simulations were performed on datasets of dimension $90 \times 20 \times 2$.
Reported values correspond to the median, with the first and third quartiles shown in parentheses.

\subsubsection{Two classes}  

\begin{figure}[h]
    \centering
    \begin{minipage}{0.24\textwidth}
        \includegraphics[width=\linewidth]{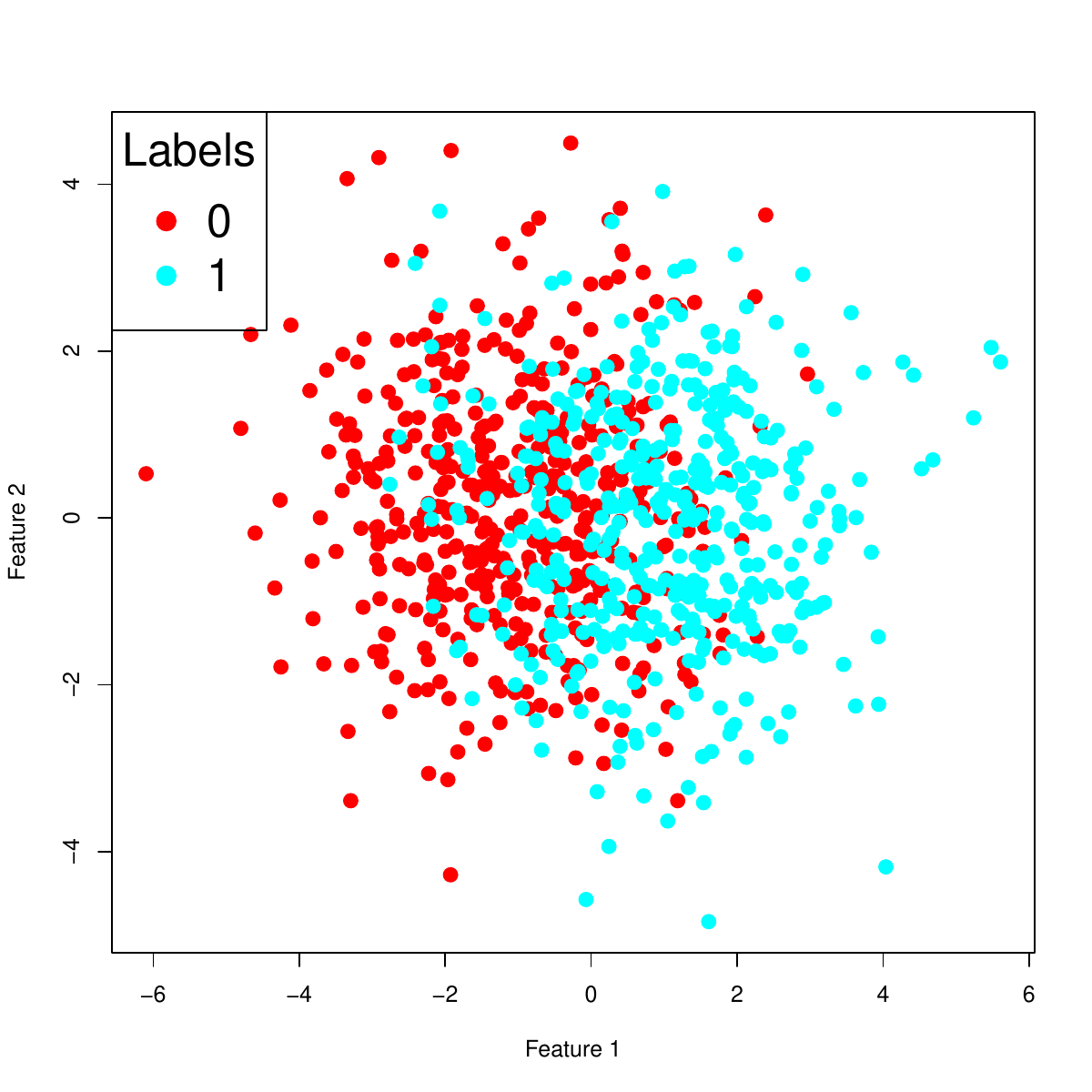}
    \end{minipage}
    \begin{minipage}{0.24\textwidth}
        \includegraphics[width=\linewidth]{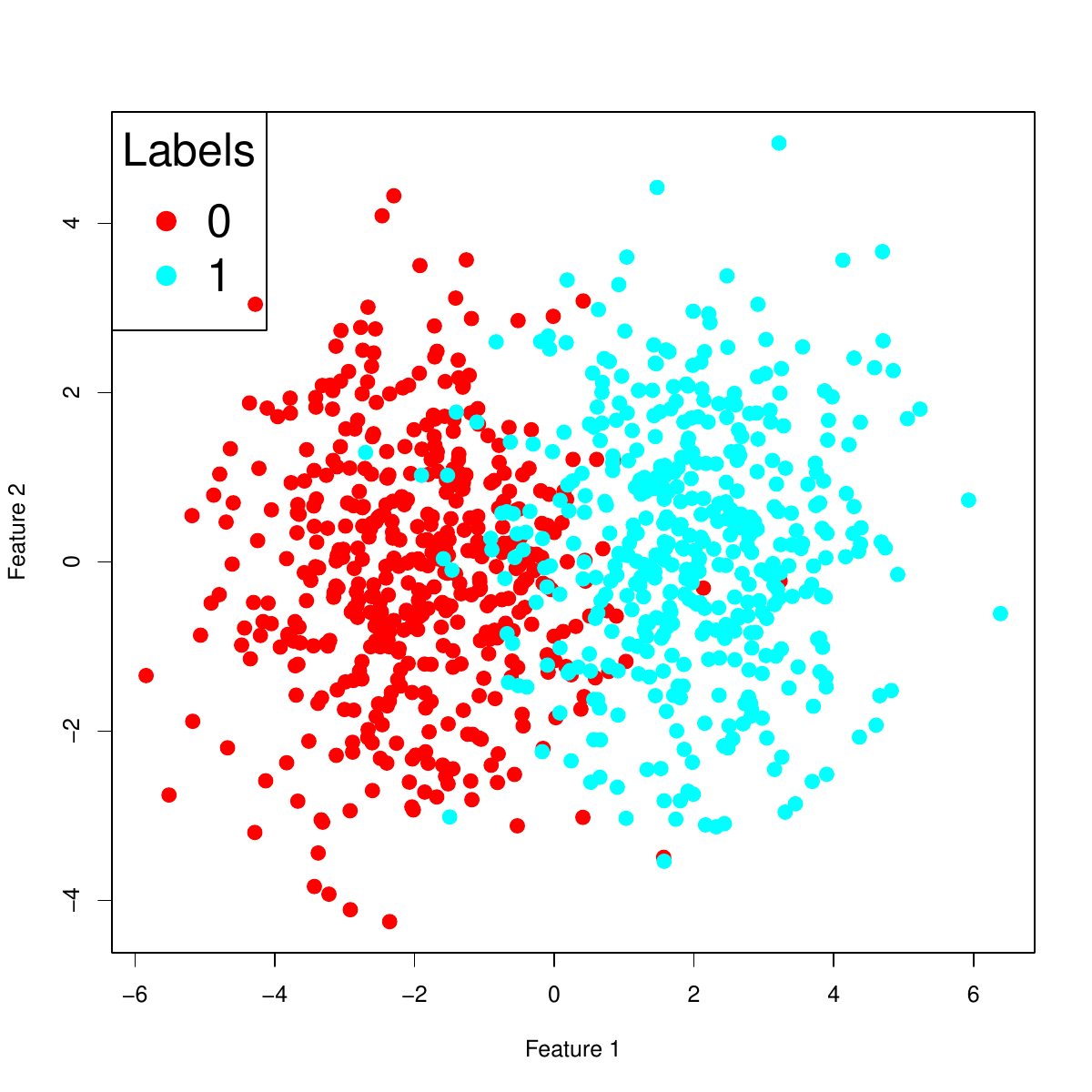}
    \end{minipage}
    \begin{minipage}{0.24\textwidth}
        \includegraphics[width=\linewidth]{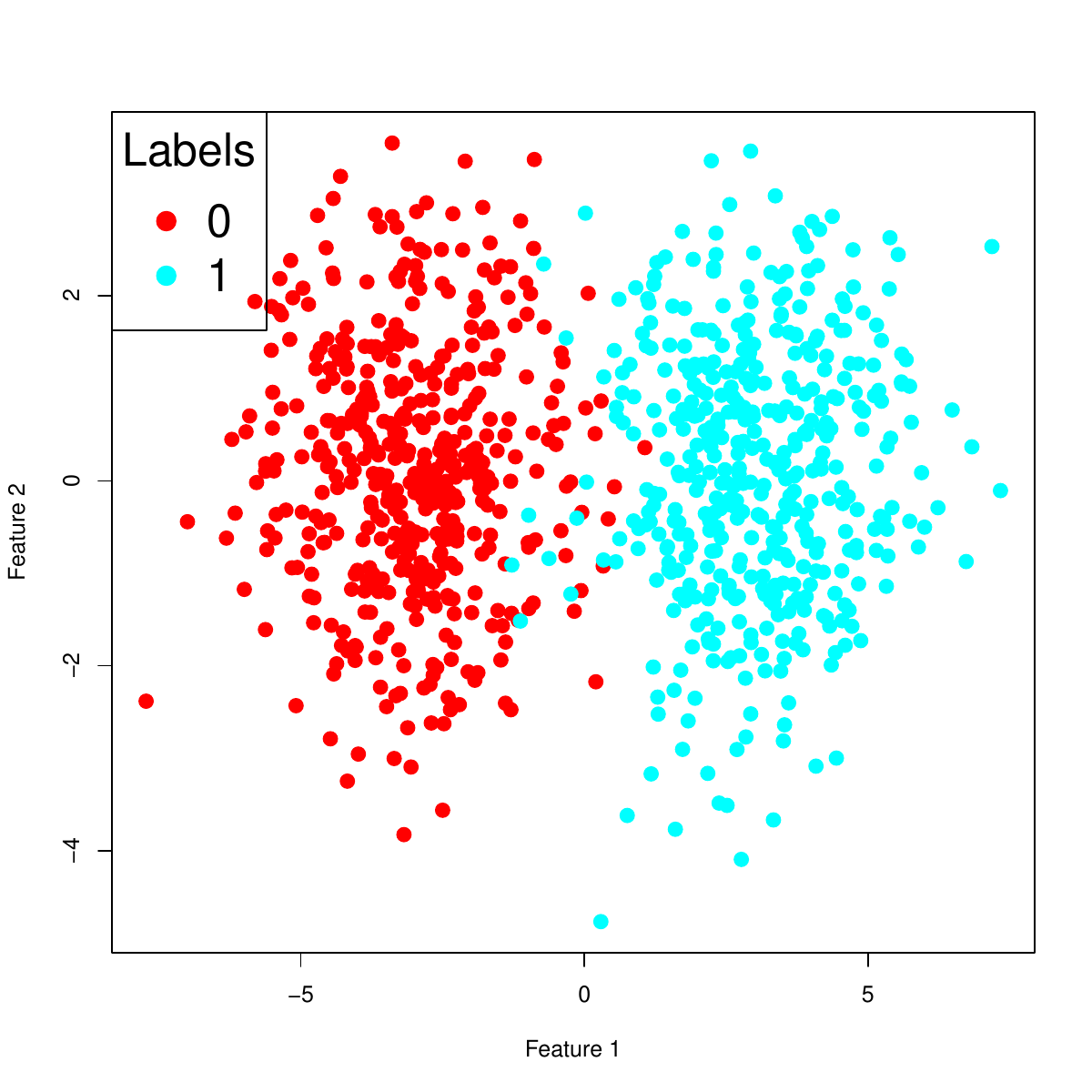}
    \end{minipage}
    \begin{minipage}{0.24\textwidth}
        \includegraphics[width=\linewidth]{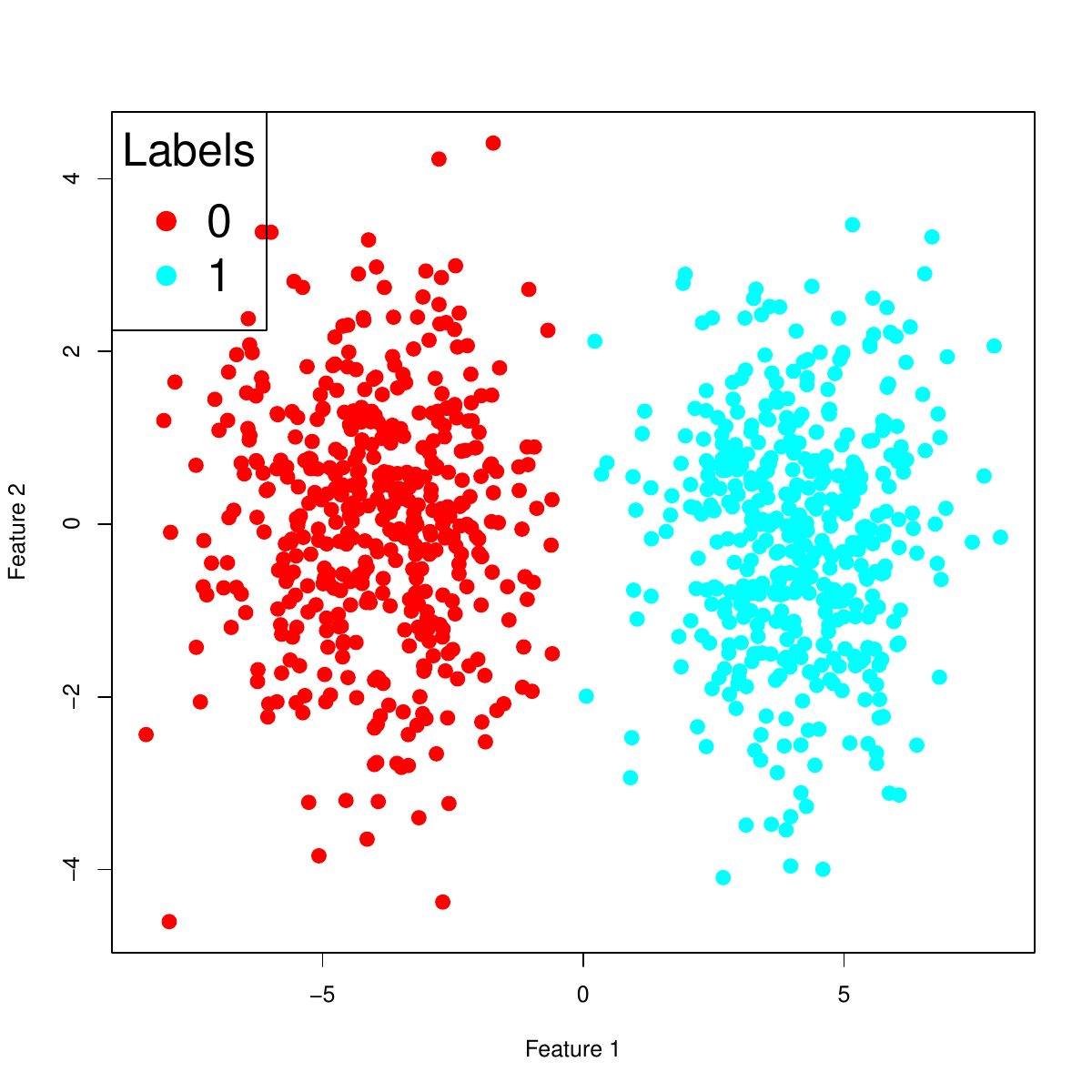}
    \end{minipage}
    \caption{Wasserstein distances for two-class point clouds with increasing separation (2, 4, 6, 8), corresponding to decreasing classification difficulty.}
    \label{fig:2 clouds difficulty}
\end{figure}

The different data settings are displayed in Figure \ref{fig:2 clouds difficulty}.
The Transformer was trained with parameters shown in table \ref{tab:params 2}, the results are shown in table \ref{tab:results_two_clouds}.

\begin{table}[h]
    \centering
    \scriptsize
    \begin{tabular}{ll}
        \toprule
        \textbf{Parameter} & \textbf{Value} \\
        \midrule
        Epochs & 200 \\
        Transformer blocks & 1 \\
        Number of heads & 10 \\
        Head dimension & 16 \\
        Feedforward dimension (SA) & 32 \\
        MLP units & 8 \\
        SA Dropout & 0.1 \\
        MLP Dropout & 0.4 \\
        Learning rate & 0.01 \\
        \bottomrule
    \end{tabular}
    \caption{Training parameters of the Transformer for 2-label classification}
    \label{tab:params 2}
\end{table}

\begin{table}[h]
    \centering
    \resizebox{\textwidth}{!}{
    \begin{tabular}{lcccc}
        \toprule
        \textbf{Wasserstein distance (between gaussians)} & \textbf{2.000} & \textbf{4.000} & \textbf{6.000} & \textbf{8.000} \\
        \midrule
        Accuracy point-wise & 1.000 (0.997–1.000) & 1.000 (1.000–1.000) & 1.000 (1.000–1.000) & 1.000 (0.999–1.000) \\
        Computational Time & 21.833 (21.670–22.170) & 22.091 (21.921–22.296) & 22.245 (21.976–22.464) & 22.356 (21.985–22.610) \\
        Matching & 1.000 (0.994–1.000) & 1.000 (0.968–1.000) & 1.000 (0.854–1.000) & 1.000 (0.930–1.000) \\
        Wasserstein Distance & 147.577 (132.722–157.646) & 129.618 (118.153–151.849) & 121.727 (103.247–163.332) & 123.880 (88.157–169.100) \\
        Transformer Distance & 147.636 (132.722–157.646) & 129.618 (119.603–155.150) & 122.051 (103.386–164.212) & 125.836 (88.726–169.127) \\
        Transformer Cost & 8581.618 (6999.051–10554.946) & 7969.820 (6626.263–11092.792) & 8164.189 (6188.536–11449.686) & 6855.794 (5424.082–9454.956) \\
        Monge Gap & 0.000 (0.000–0.002) & 0.000 (0.000–0.059) & 0.000 (0.000–0.376) & 0.000 (0.000–0.233) \\
        Efficiency & 0.016 (0.013–0.022) & 0.016 (0.012–0.021) & 0.016 (0.011–0.022) & 0.017 (0.012–0.024) \\
        Optimality & 1.000 (1.000–1.000) & 1.000 (0.999–1.000) & 1.000 (0.997–1.000) & 1.000 (0.998–1.000) \\
        Best Epoch & 197.000 (181.250–199.000) & 197.000 (184.750–199.000) & 197.000 (191.750–199.000) & 198.000 (190.000–199.000) \\
        Accuracy instance-wise & 1.000 (1.000–1.000) & 1.000 (1.000–1.000) & 1.000 (1.000–1.000) & 1.000 (1.000–1.000) \\
        Recall Class 0 & 1.000 (1.000–1.000) & 1.000 (1.000–1.000) & 1.000 (1.000–1.000) & 1.000 (1.000–1.000) \\
        Recall Class 1 & 1.000 (1.000–1.000) & 1.000 (1.000–1.000) & 1.000 (1.000–1.000) & 1.000 (1.000–1.000) \\
        \bottomrule
    \end{tabular}}
    \caption{Summary of results for two-label classification with Transformer algorithm. Values are reported as median (first-third quartile).}
    \label{tab:results_two_clouds}
\end{table}

\subsubsection{Three classes}  

\begin{figure}[h]
    \centering
    \begin{minipage}{0.24\textwidth}
        \includegraphics[width=\linewidth]{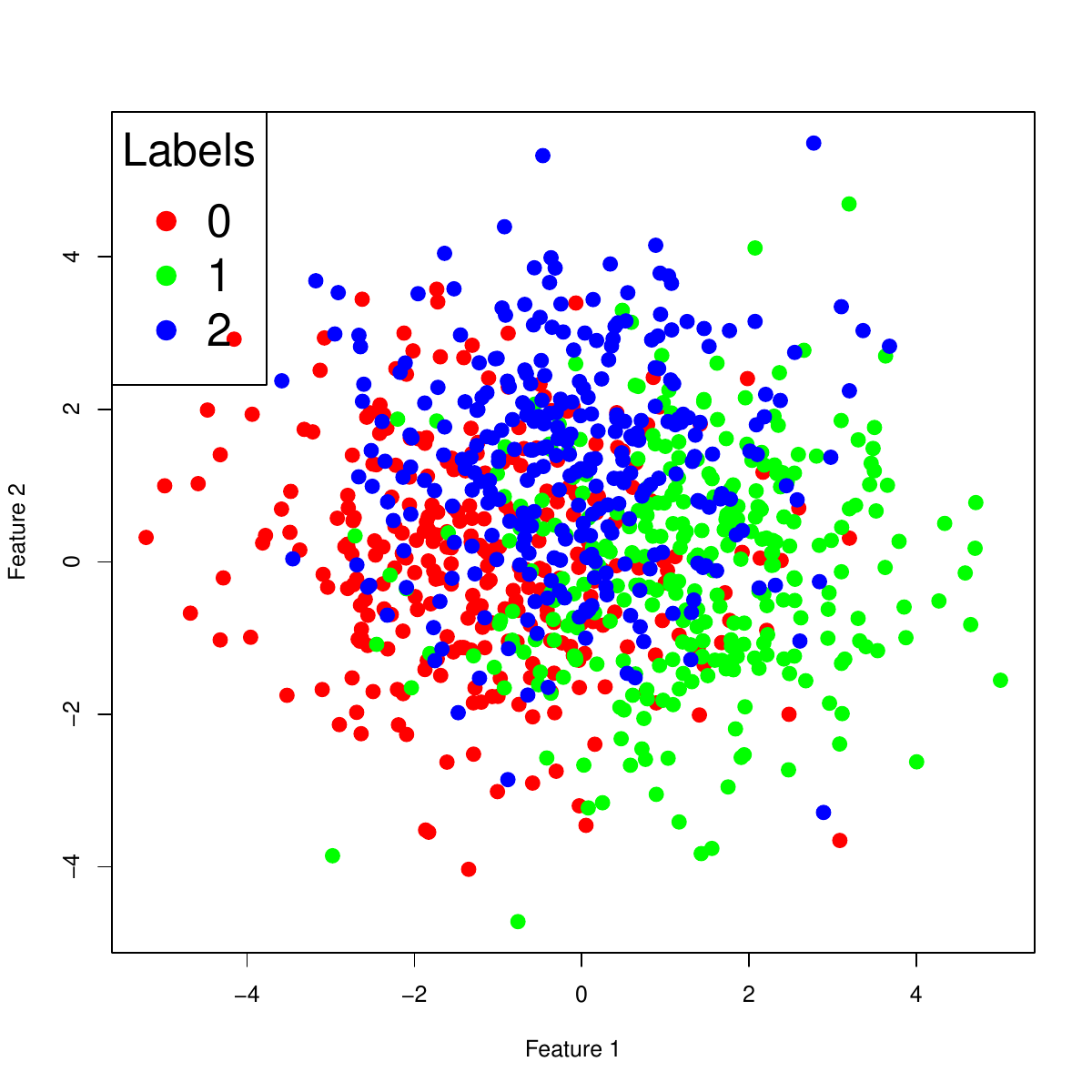}
    \end{minipage}
    \begin{minipage}{0.24\textwidth}
        \includegraphics[width=\linewidth]{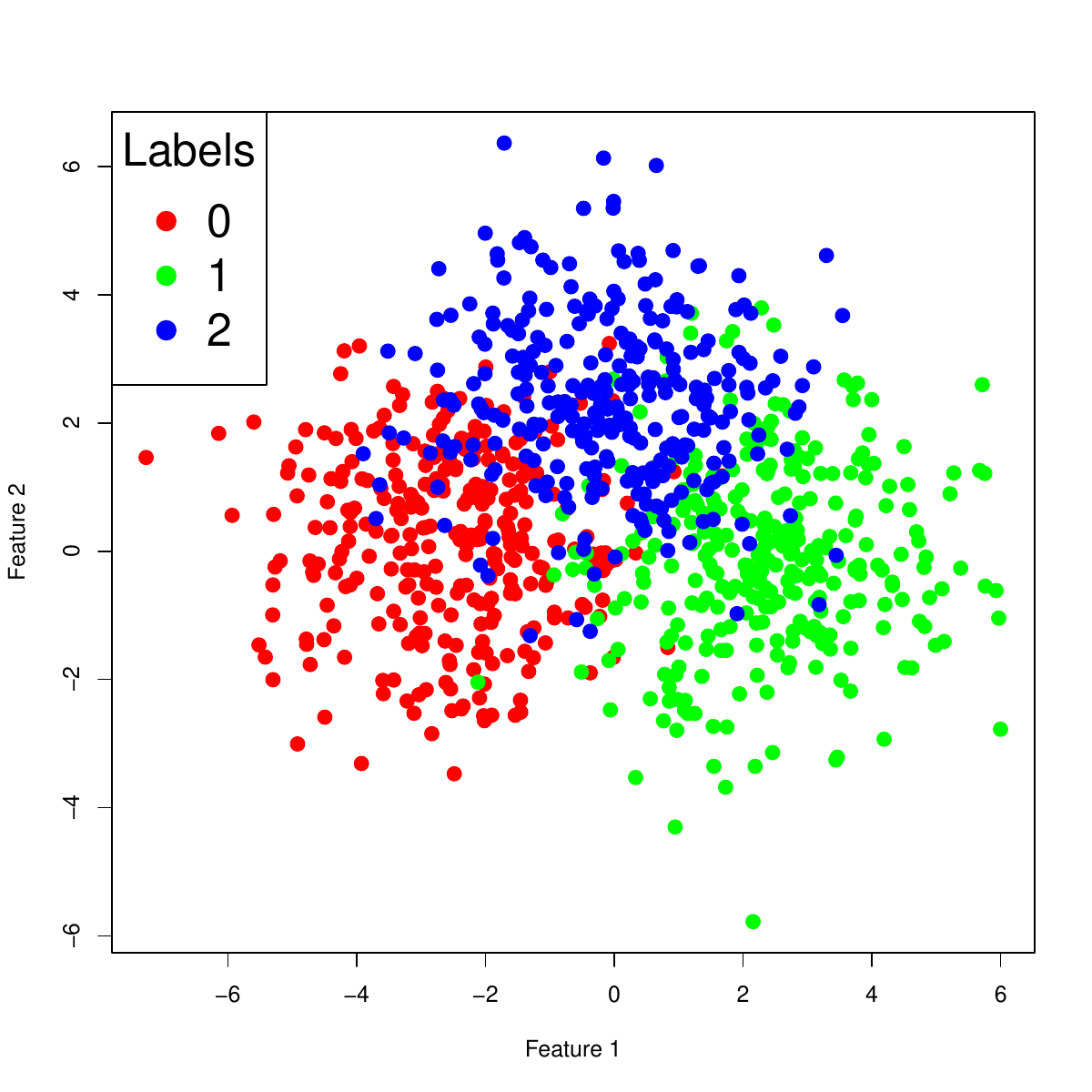}
    \end{minipage}
    \begin{minipage}{0.24\textwidth}
        \includegraphics[width=\linewidth]{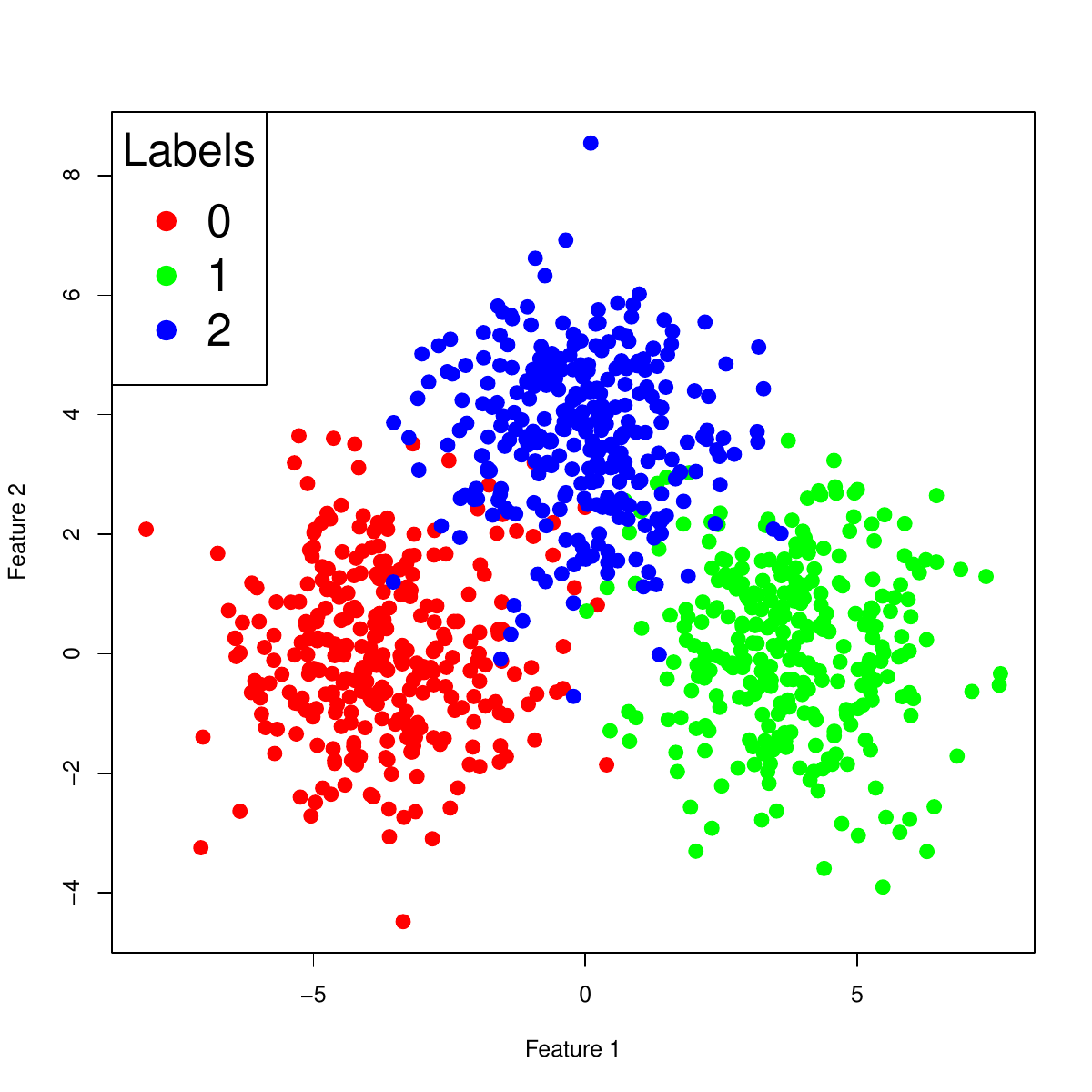}
    \end{minipage}
    \begin{minipage}{0.24\textwidth}
        \includegraphics[width=\linewidth]{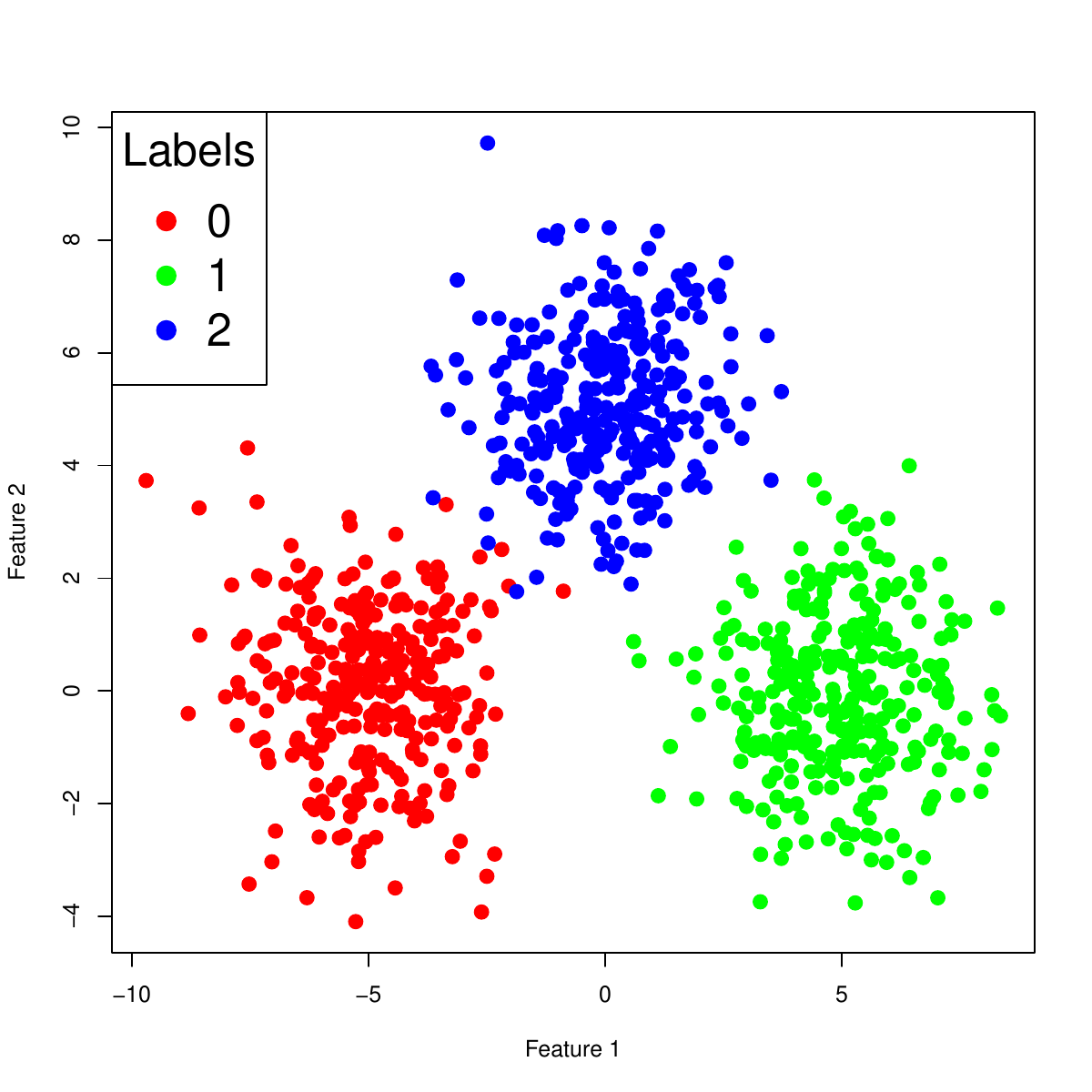}
    \end{minipage}
    \caption{Average Wasserstein distances for three-class point clouds with increasing separation (2, 4, 6, 8), corresponding to decreasing classification difficulty.}
    \label{fig:3 clouds difficulty}
\end{figure}

The different data settings are displayed in Figure \ref{fig:3 clouds difficulty}.
The Transformer was trained with parameters shown in table \ref{tab:params 3}, the results are shown in table \ref{tab:results_three_clouds_summary}.

\begin{table}[h]
    \centering
    \scriptsize
    \begin{tabular}{ll}
        \toprule
        \textbf{Parameter} & \textbf{Value} \\
        \midrule
        Epochs & 200 \\
        Transformer blocks & 2 \\
        Number of heads & 10 \\
        Head dimension & 16 \\
        Feedforward dimension (SA) & 32 \\
        MLP units & 8 \\
        SA Dropout & 0.1 \\
        MLP Dropout & 0.4 \\
        Learning rate & 0.01 \\
        \bottomrule
    \end{tabular}
    \caption{Training parameters of the Transformer for 3-label classification}
    \label{tab:params 3}
\end{table}

\begin{table}[h]
    \centering
    \resizebox{\textwidth}{!}{
    \begin{tabular}{lcccc}
        \toprule
        \textbf{mean Wasserstein distance (between gaussians)} & \textbf{2.000} & \textbf{4.000} & \textbf{6.000} & \textbf{8.000} \\
        \midrule
        Accuracy point-wise & 0.808 (0.786–0.866) & 0.910 (0.857–0.926) & 0.965 (0.929–0.985) & 0.973 (0.912–0.990) \\
        Computational Time & 38.610 (35.725–41.048) & 47.458 (42.521–52.457) & 64.262 (60.483–68.062) & 69.979 (64.867–75.154) \\
        Matching & 0.668 (0.404–0.943) & 0.727 (0.501–0.983) & 0.796 (0.656–1.000) & 0.792 (0.672–0.965) \\
        Wasserstein Distance & 192.785 (165.527–232.905) & 168.168 (136.132–200.255) & 147.900 (117.514–169.895) & 132.065 (102.381–170.056) \\
        Transformer Distance & 196.452 (168.666–234.622) & 169.764 (138.030–202.255) & 148.421 (118.575–171.661) & 132.341 (106.486–174.657) \\
        Transformer Cost & 22283.637 (17079.808–29911.348) & 19258.328 (16433.342–24763.441) & 18782.371 (15734.530–21589.795) & 17539.566 (14210.767–20233.784) \\
        Monge Gap & 2.131 (0.094–6.082) & 1.089 (0.012–3.398) & 0.399 (0.000–2.714) & 0.423 (0.016–2.444) \\
        Efficiency & 0.009 (0.006–0.011) & 0.009 (0.007–0.011) & 0.008 (0.007–0.010) & 0.008 (0.006–0.011) \\
        Optimality & 0.989 (0.971–1.000) & 0.993 (0.980–1.000) & 0.997 (0.980–1.000) & 0.997 (0.981–1.000) \\
        Best Epoch & 172.500 (120.750–190.250) & 170.000 (122.750–195.000) & 178.000 (140.250–194.000) & 180.500 (122.500–195.000) \\
        Accuracy instance-wise & 0.978 (0.933–1.000) & 1.000 (0.978–1.000) & 1.000 (1.000–1.000) & 1.000 (1.000–1.000) \\
        Recall Class 0 & 1.000 (1.000–1.000) & 1.000 (1.000–1.000) & 1.000 (1.000–1.000) & 1.000 (1.000–1.000) \\
        Recall Class 1 & 1.000 (1.000–1.000) & 1.000 (1.000–1.000) & 1.000 (1.000–1.000) & 1.000 (1.000–1.000) \\
        Recall Class 2 & 1.000 (0.933–1.000) & 1.000 (1.000–1.000) & 1.000 (1.000–1.000) & 1.000 (1.000–1.000) \\
                \bottomrule
    \end{tabular}}
    \caption{Summary of results for three-label classification with Transformer algorithm. Values are reported as median (first-third quartile).}
    \label{tab:results_three_clouds_summary}
\end{table}

As expected, the Transformer achieves high performance.  
From the \textbf{matching} metric we observe that the SA remapping is largely suboptimal in the three-label setup, which worsens as the classification task becomes more difficult.
However, this does not translate into markedly low values of \textbf{Optimality} or high values of \textbf{Monge Gap}.  
Given that the \textbf{Efficiency} measure remains very low, we conclude that SA reaches a solution that is reasonably close to the optimal and efficient one, but the trajectory it follows to get there is highly inefficient.
A possible explanation is that at each training iteration gradient descent simultaneously optimizes both the SA blocks (which performs the remapping) and the MLP block (which carries out the classification).
This joint optimization may introduce inefficiencies.
To investigate this, I propose first training the MLP block on dummy data that are easy to classify, then freezing it, and finally training the SA block on the real data.

\subsection{pretrained Transformer}

The idea behind the pretrained Transformer is to disentangle the roles of the two components of the architecture: the MLP block, responsible for classification, and the SA block, responsible for remapping the input data.
Instead of training them jointly, the MLP is first trained on \textit{dummy data} that are simple to classify. Once this block has converged, it is frozen, and only then is the SA block trained on the \textit{real data}.

\begin{figure}[h]
    \centering
    \begin{subfigure}[b]{0.45\linewidth}
        \centering
        \includegraphics[width=\linewidth]{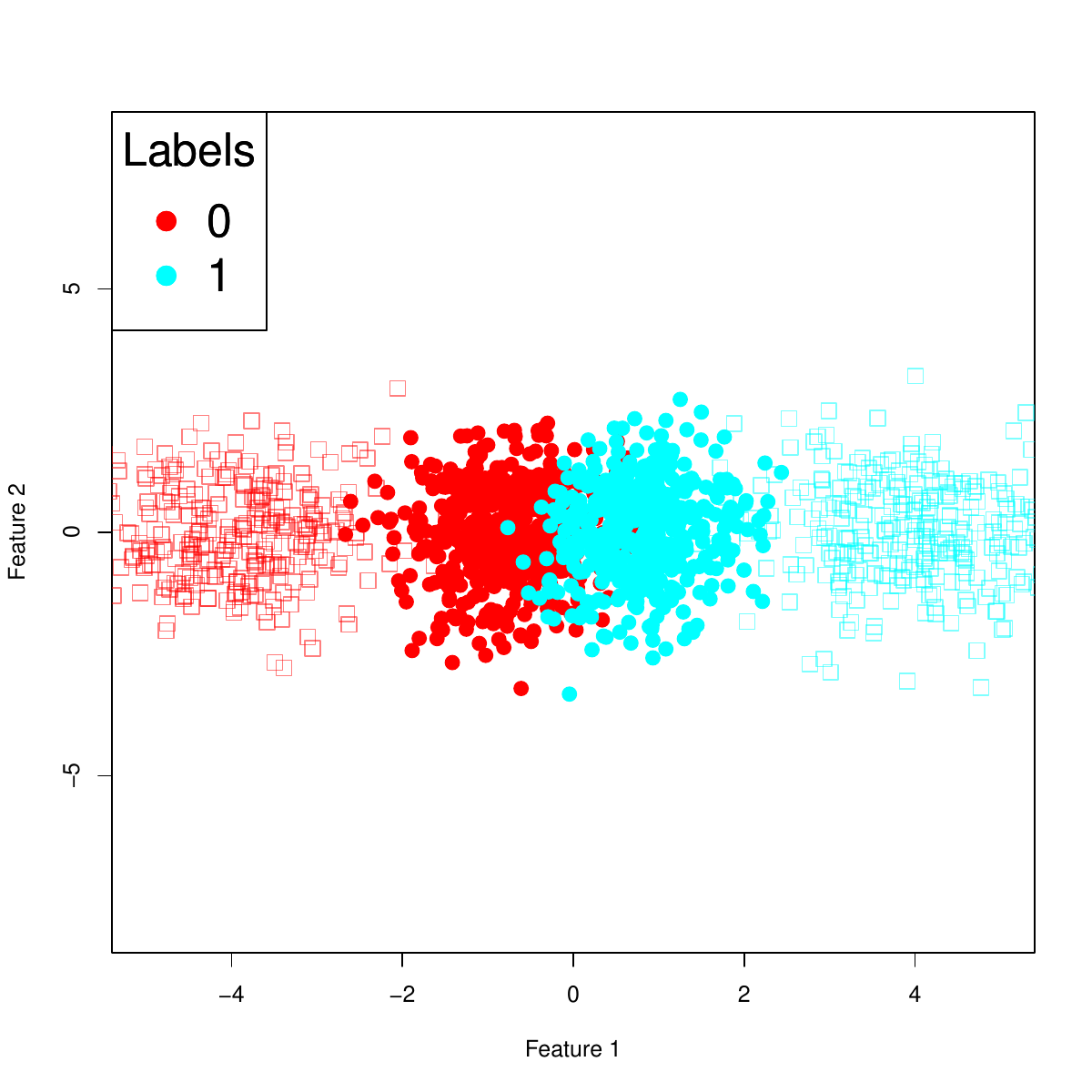}
    \end{subfigure}
    \hfill
    \begin{subfigure}[b]{0.45\linewidth}
        \centering
        \includegraphics[width=\linewidth]{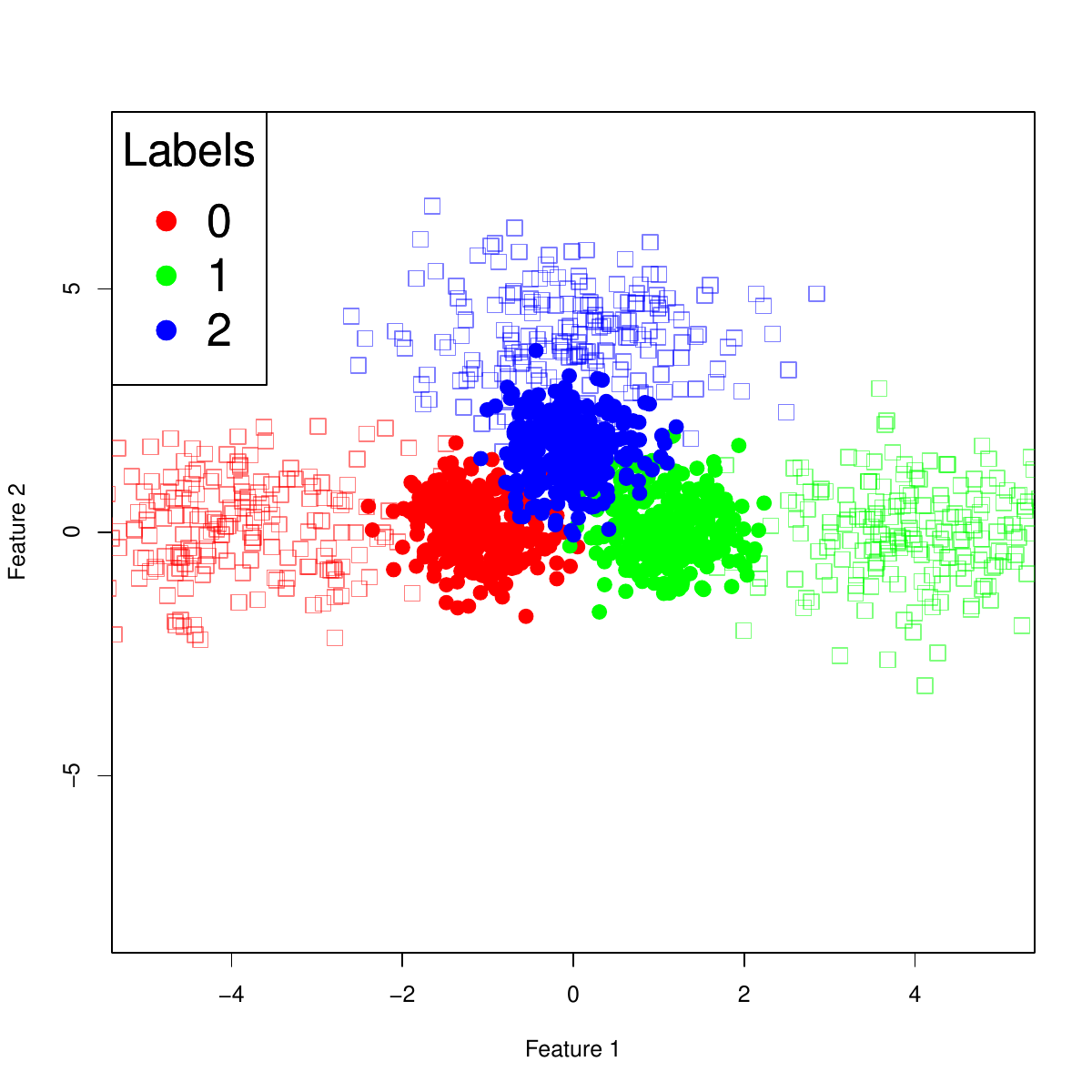}
    \end{subfigure}
    \caption{Initialization of the dummy data (empty squares) with respect to the real data (full points).}
    \label{fig:iniz}
\end{figure}

The \textit{dummy data} was initialized as shown in Figure \ref{fig:iniz} which is to say in the most favorable way, the training parameters are shown in table \ref{tab: params pretrained}, the results are shown in tables \ref{tab:results_two_clouds_pretrained_0}
and \ref{tab:results_three_clouds_pretrained_0}.

\begin{table}[h]
    \centering
    \scriptsize
    \begin{tabular}{ll}
        \toprule
        \textbf{Parameter} & \textbf{Value} \\
        \midrule
        Epochs & 200 \\
        Transformer blocks & 1 \\
        Number of heads & 10 \\
        Head dimension & 16 \\
        Feedforward dimension (SA) & 32 \\
        MLP units & 8 \\
        SA Dropout & 0.1 \\
        MLP Dropout & 0.4 \\
        Learning rate & 0.01 \\
        \bottomrule
    \end{tabular}
    \caption{Training parameters of the pretrained Transformer for 2-label classification}
    \label{tab: params pretrained}
\end{table}

\begin{table}[h]
    \centering
    \resizebox{\textwidth}{!}{
    \begin{tabular}{lcccc}
        \toprule
        \textbf{Wasserstein distance (between gaussians)} & \textbf{2.000} & \textbf{4.000} & \textbf{6.000} & \textbf{8.000} \\
        \midrule
        Accuracy point-wise & 1.000 (0.999–1.000) & 1.000 (1.000–1.000) & 1.000 (1.000–1.000) & 1.000 (1.000–1.000) \\
        Computational Time & 22.182 (21.950–22.430) & 22.323 (22.096–22.647) & 22.465 (22.240–22.777) & 23.179 (22.785–23.600) \\
        Matching & 1.000 (0.998–1.000) & 1.000 (1.000–1.000) & 1.000 (1.000–1.000) & 1.000 (1.000–1.000) \\
        Wasserstein Distance & 124.370 (114.733–136.186) & 125.672 (118.579–135.939) & 147.205 (121.066–159.349) & 156.917 (110.125–173.138) \\
        Transformer Distance & 124.370 (114.989–136.186) & 125.882 (118.618–136.060) & 147.205 (121.501–159.349) & 156.917 (110.125–173.138) \\
        Transformer Cost & 7443.061 (6602.329–8767.528) & 6621.234 (6064.807–8159.074) & 6301.713 (5890.336–7064.905) & 5423.829 (5206.628–6348.798) \\
        Monge Gap & 0.000 (0.000–0.000) & 0.000 (0.000–0.000) & 0.000 (0.000–0.000) & 0.000 (0.000–0.000) \\
        Efficiency & 0.016 (0.014–0.019) & 0.019 (0.015–0.022) & 0.023 (0.017–0.027) & 0.030 (0.019–0.033) \\
        Optimality & 1.000 (1.000–1.000) & 1.000 (1.000–1.000) & 1.000 (1.000–1.000) & 1.000 (1.000–1.000) \\
        Best Epoch & 198.000 (171.000–199.000) & 199.000 (196.000–199.000) & 199.000 (198.000–199.000) & 199.000 (199.000–199.000) \\
        Accuracy instance-wise & 1.000 (1.000–1.000) & 1.000 (1.000–1.000) & 1.000 (1.000–1.000) & 1.000 (1.000–1.000) \\
        Recall Class 0 & 1.000 (1.000–1.000) & 1.000 (1.000–1.000) & 1.000 (1.000–1.000) & 1.000 (1.000–1.000) \\
        Recall Class 1 & 1.000 (1.000–1.000) & 1.000 (1.000–1.000) & 1.000 (1.000–1.000) & 1.000 (1.000–1.000) \\
        \bottomrule
    \end{tabular}}
    \caption{Summary of results for two-label classification with pretrained Transformer (aligned \ref{fig:init_0}). Values are reported as median (first–third quartile).}
    \label{tab:results_two_clouds_pretrained_0}
\end{table}

\clearpage

\begin{table}[h]
    \centering
    \scriptsize
    \begin{tabular}{ll}
        \toprule
        \textbf{Parameter} & \textbf{Value} \\
        \midrule
        Epochs & 200 \\
        Transformer blocks & 2 \\
        Number of heads & 10 \\
        Head dimension & 16 \\
        Feedforward dimension (SA) & 32 \\
        MLP units & 8 \\
        SA Dropout & 0.1 \\
        MLP Dropout & 0.4 \\
        Learning rate & 0.01 \\
        \bottomrule
    \end{tabular}
    \caption{Training parameters of the pretrained Transformer for 3-label classification}
    \label{tab:params 3 pre}
\end{table}
    
\begin{table}[h]
    \centering
    \resizebox{\textwidth}{!}{
    \begin{tabular}{lcccc}
        \toprule
        \textbf{mean Wasserstein distance (between gaussians)} & \textbf{2.000} & \textbf{4.000} & \textbf{6.000} & \textbf{8.000} \\
        \midrule
        Accuracy point-wise & 0.802 (0.772–0.926) & 0.922 (0.906–0.994) & 0.971 (0.951–0.994) & 0.986 (0.953–0.994) \\
        Computational Time & 25.374 (25.278–25.552) & 25.350 (25.129–25.557) & 25.488 (25.250–25.776) & 25.740 (25.573–26.448) \\
        Matching & 0.511 (0.355–0.784) & 0.762 (0.558–0.997) & 0.774 (0.659–0.997) & 0.765 (0.673–0.894) \\
        Wasserstein Distance & 114.530 (97.297–127.055) & 141.211 (118.092–182.549) & 131.971 (103.729–169.977) & 124.887 (102.618–149.107) \\
        Transformer Distance & 120.574 (103.836–134.079) & 144.181 (121.604–184.557) & 132.922 (104.922–170.684) & 125.639 (103.058–151.290) \\
        Transformer Cost & 18350.224 (11424.580–24013.723) & 14638.747 (8093.737–19670.762) & 15422.579 (10505.070–19736.344) & 14123.791 (8701.208–19294.373) \\
        Monge Gap & 4.798 (1.802–8.664) & 1.379 (0.004–3.659) & 0.640 (0.013–2.018) & 0.636 (0.090–2.119) \\
        Efficiency & 0.006 (0.005–0.010) & 0.010 (0.007–0.018) & 0.009 (0.007–0.014) & 0.009 (0.007–0.013) \\
        Optimality & 0.957 (0.925–0.986) & 0.992 (0.974–1.000) & 0.996 (0.986–1.000) & 0.996 (0.984–0.999) \\
        Best Epoch & 139.000 (86.000–189.000) & 133.000 (41.750–187.500) & 130.000 (79.750–178.250) & 126.000 (48.750–184.500) \\
        Accuracy instance-wise & 0.978 (0.911–1.000) & 1.000 (1.000–1.000) & 1.000 (1.000–1.000) & 1.000 (1.000–1.000) \\
        Recall Class 0 & 1.000 (1.000–1.000) & 1.000 (1.000–1.000) & 1.000 (1.000–1.000) & 1.000 (1.000–1.000) \\
        Recall Class 1 & 1.000 (1.000–1.000) & 1.000 (1.000–1.000) & 1.000 (1.000–1.000) & 1.000 (1.000–1.000) \\
        Recall Class 2 & 0.933 (0.733–1.000) & 1.000 (1.000–1.000) & 1.000 (1.000–1.000) & 1.000 (1.000–1.000) \\
        \bottomrule
    \end{tabular}}
    \caption{Summary of results for three-label classification with pretrained Transformer (aligned \ref{fig:init_three_0}). Values are reported as median (first–third quartile).}
    \label{tab:results_three_clouds_pretrained_0}
\end{table}

\begin{figure}[h]
    \centering
    \begin{subfigure}[b]{0.3\linewidth}
        \centering
        \includegraphics[width=\linewidth]{plots/inzializzazione_2_0.pdf}
        \subcaption{aligned}
        \label{fig:init_0}
    \end{subfigure}
    \hfill
    \begin{subfigure}[b]{0.3\linewidth}
        \centering
        \includegraphics[width=\linewidth]{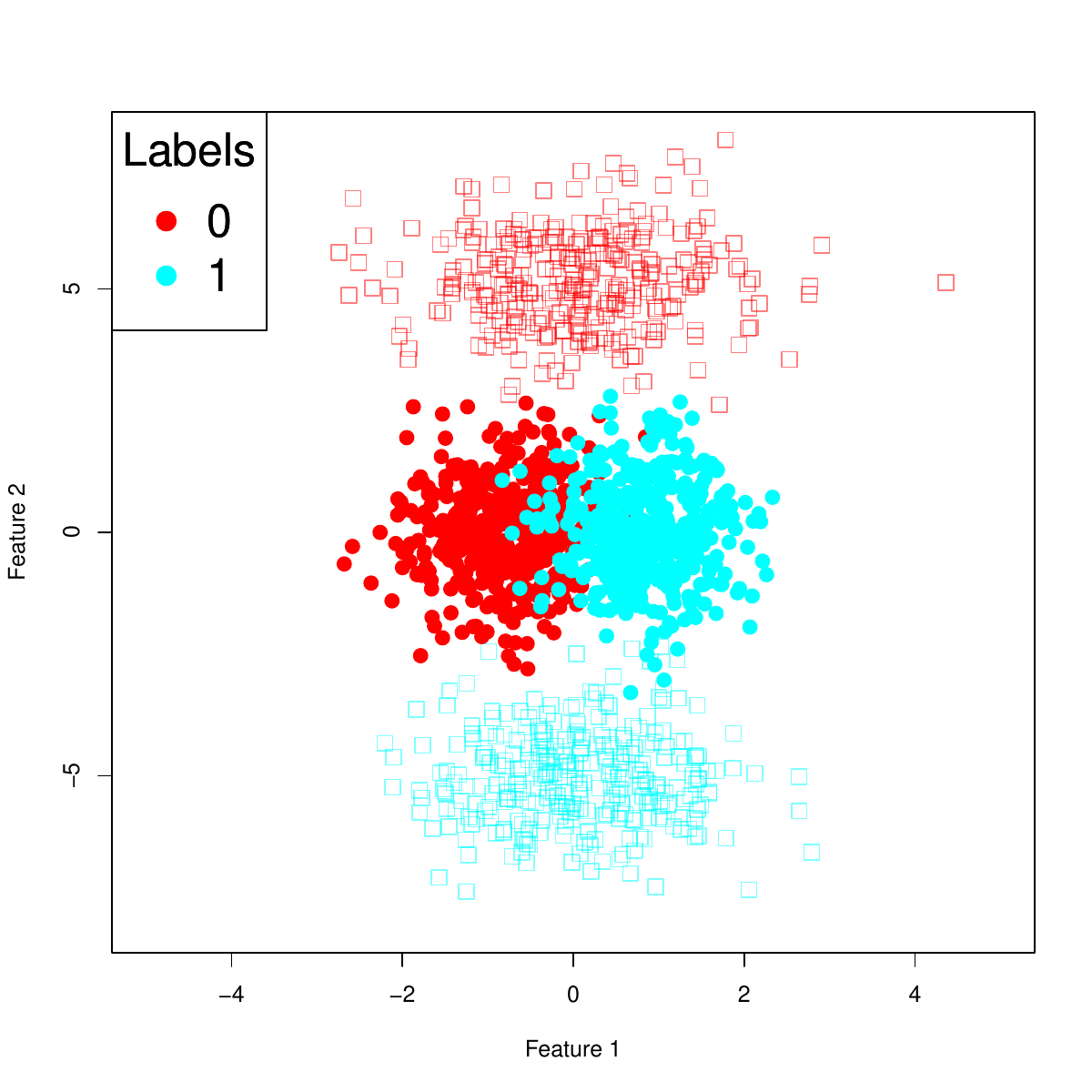}
        \subcaption{$90^\circ$ rotation}
        \label{fig:init_90}
    \end{subfigure}
    \hfill
    \begin{subfigure}[b]{0.3\linewidth}
        \centering
        \includegraphics[width=\linewidth]{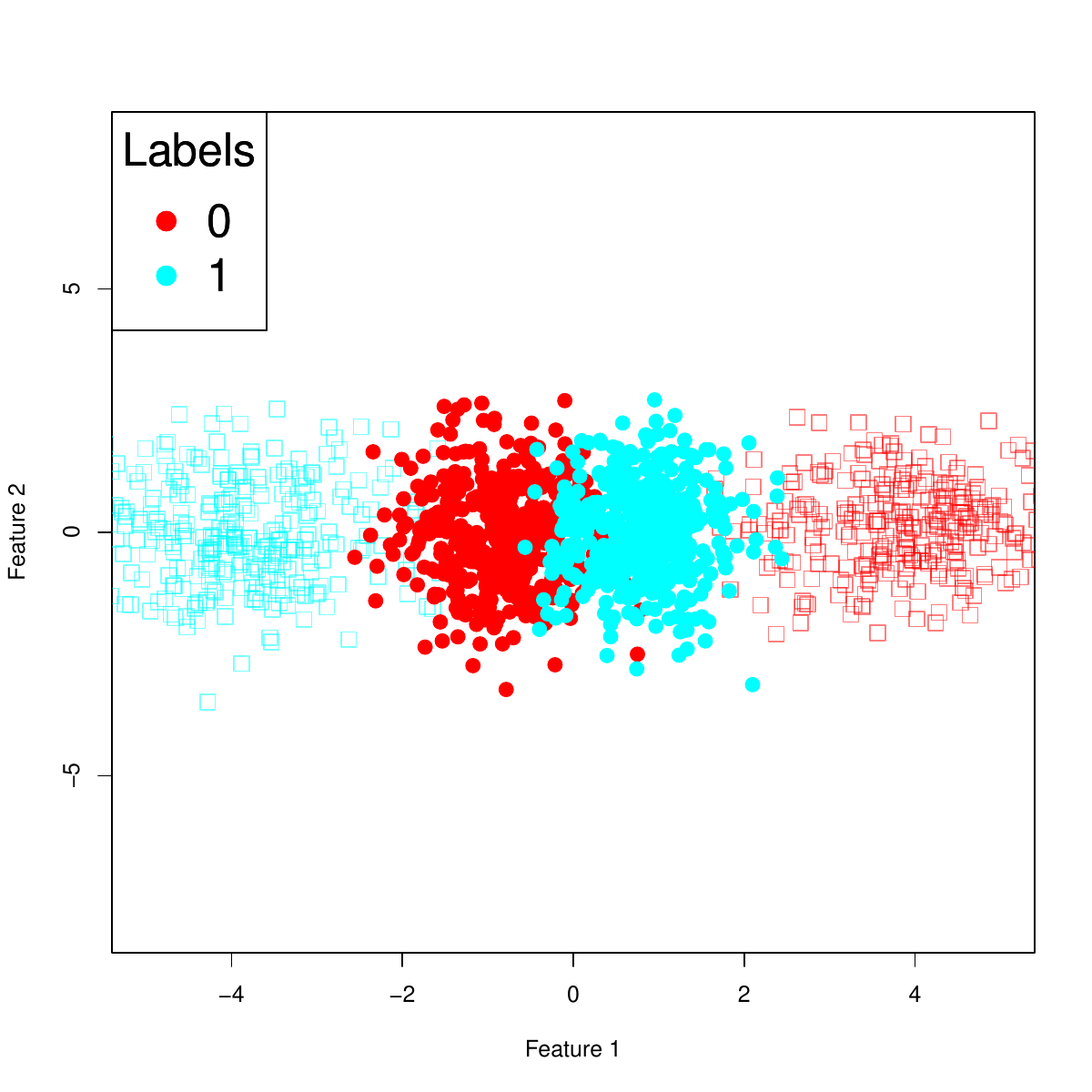}
        \subcaption{$180^\circ$ rotation}
        \label{fig:init_180}
    \end{subfigure}
    \caption{Initialization of dummy data under three different orientations with respect to the real data. 
    The subfigures show the favorable alignment and the rotated configurations ($90^\circ$, $180^\circ$) used to test sensitivity to initialization. The full points are the \textit{real data} and the empty squares are \textit{dummy data}.}
    \label{fig:dummy_inits}
\end{figure}

\begin{figure}[h]
    \centering
    \begin{subfigure}[b]{0.3\linewidth}
        \centering
        \includegraphics[width=\linewidth]{plots/inzializzazione_3_0.pdf}
        \subcaption{aligned}
        \label{fig:init_three_0}
    \end{subfigure}
    \hfill
    \begin{subfigure}[b]{0.3\linewidth}
        \centering
        \includegraphics[width=\linewidth]{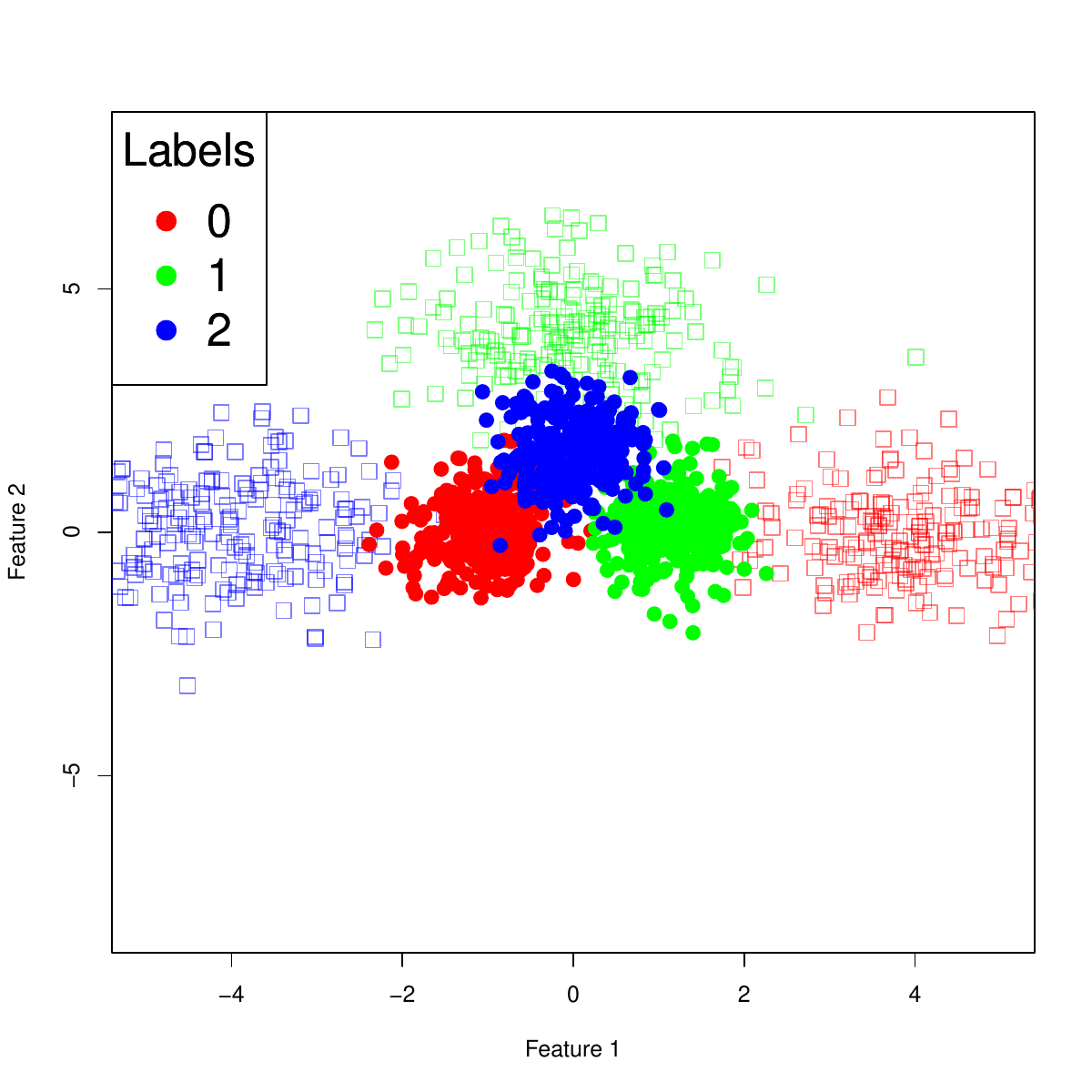}
        \subcaption{$120^\circ$ rotation}
        \label{fig:init_three_120}
    \end{subfigure}
    \hfill
    \begin{subfigure}[b]{0.3\linewidth}
        \centering
        \includegraphics[width=\linewidth]{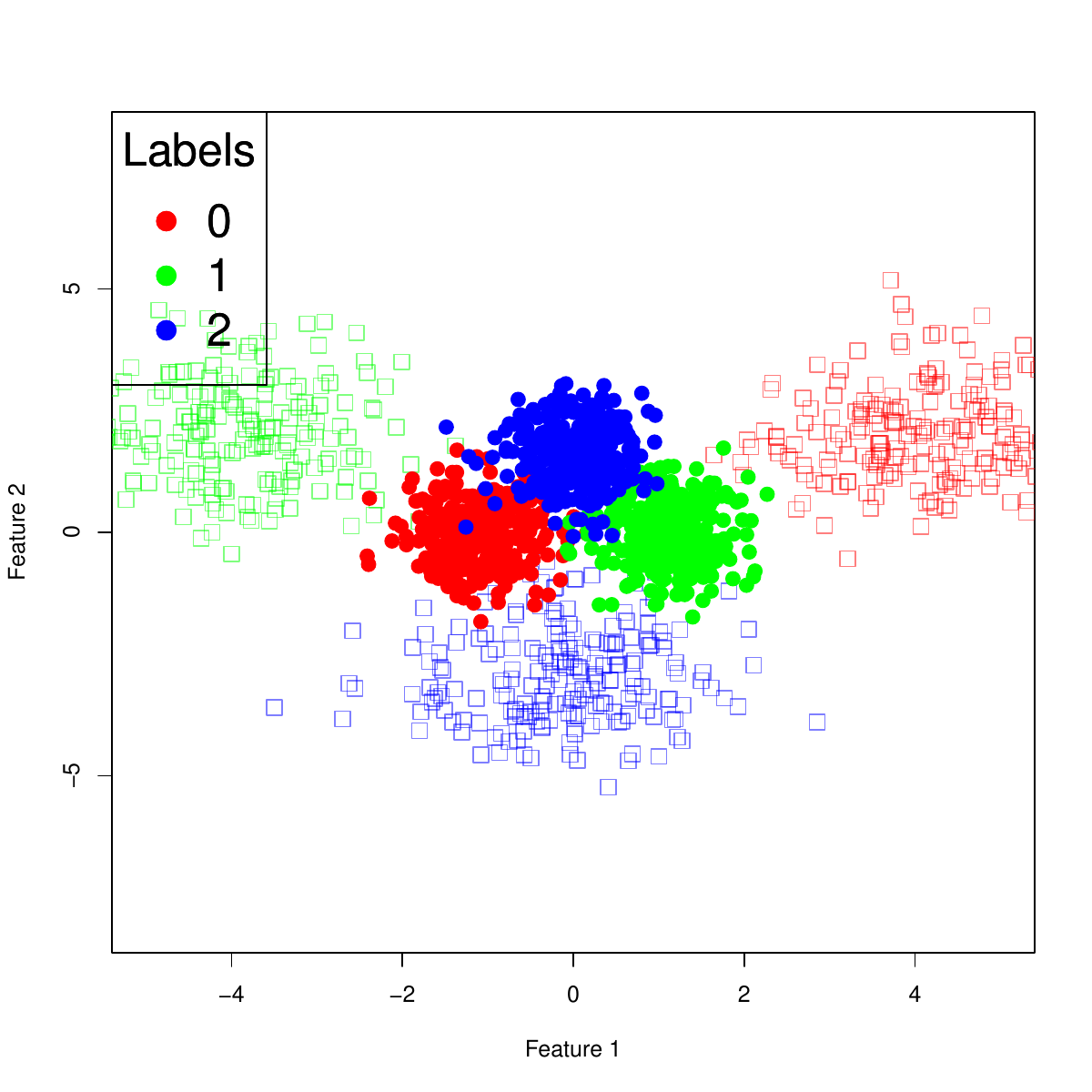}
        \subcaption{$180^\circ$ rotation}
        \label{fig:init_three_180}
    \end{subfigure}
    \caption{Initialization of dummy data for the three-clouds setup under different orientations. 
    The subfigures show the favorable alignment and the rotated configurations ($120^\circ$, $180^\circ$) used to test the sensitivity of the model to initialization. 
    The full points correspond to the \textit{real data}, while the empty squares represent the \textit{dummy data}.}
    \label{fig:dummy_inits_three}
\end{figure}

This setup improves \textbf{Accuracy} and \textbf{Matching}, while the improvement in \textbf{Efficiency} is very small and, in truth, somewhat questionable. 
Overall, these results suggest that pre-training the MLP block gives the model a slight advantage, but it does not resolve the broader inefficiency of the approach.
An additional point concerns the initialization of the \textit{dummy data}. 
In the default setting (Figures \ref{fig:iniz}), the dummy points were placed in the most favorable configuration. 
To test the sensibility of the algorithm to this initialization I then rotated the orientation of the dummy data from the first configuration and reran the simulations, Figures \ref{fig:dummy_inits} and \ref{fig:dummy_inits_three} illustrate the setups that were considered, tables \ref{tab:results_two_clouds_pretrained_90}, \ref{tab:results_two_clouds_pretrained_180}, \ref{tab:results_three_clouds_pretrained_120},
\ref{tab:results_three_clouds_pretrained_180}  report the results.

\begin{table}[h]
    \centering
    \resizebox{\textwidth}{!}{
    \begin{tabular}{lcccc}
        \toprule
        \textbf{Wasserstein distance (between gaussians)} & \textbf{2.000} & \textbf{4.000} & \textbf{6.000} & \textbf{8.000} \\
        \midrule
        Accuracy point-wise & 0.999 (0.988–1.000) & 1.000 (1.000–1.000) & 1.000 (1.000–1.000) & 1.000 (0.999–1.000) \\
        Computational Time & 24.728 (24.485–25.048) & 24.613 (24.511–24.733) & 24.756 (24.649–24.931) & 25.261 (24.810–26.118) \\
        Matching & 0.963 (0.767–1.000) & 1.000 (1.000–1.000) & 1.000 (1.000–1.000) & 1.000 (0.942–1.000) \\
        Wass Distance & 161.835 (144.409–173.747) & 200.730 (192.853–208.851) & 208.891 (190.454–218.020) & 181.121 (158.684–212.473) \\
        Transformer Distance & 162.234 (144.719–173.747) & 200.730 (192.853–208.851) & 208.911 (190.454–218.020) & 181.121 (158.684–212.617) \\
        Transformer Cost & 7418.296 (6568.354–8645.481) & 6951.255 (6559.817–7619.394) & 7164.319 (6827.896–8059.798) & 7402.120 (6314.910–8588.444) \\
        Monge Gap & 0.054 (0.000–0.406) & 0.000 (0.000–0.000) & 0.000 (0.000–0.000) & 0.000 (0.000–0.107) \\
        Efficiency & 0.021 (0.018–0.025) & 0.030 (0.025–0.031) & 0.029 (0.022–0.031) & 0.027 (0.020–0.030) \\
        Optimality & 1.000 (0.997–1.000) & 1.000 (1.000–1.000) & 1.000 (1.000–1.000) & 1.000 (0.999–1.000) \\
        Best Epoch & 195.000 (139.750–199.000) & 199.000 (198.000–199.000) & 199.000 (198.000–199.000) & 183.500 (134.500–199.000) \\
        Accuracy instance-wise & 1.000 (1.000–1.000) & 1.000 (1.000–1.000) & 1.000 (1.000–1.000) & 1.000 (1.000–1.000) \\
        Recall Class 0 & 1.000 (1.000–1.000) & 1.000 (1.000–1.000) & 1.000 (1.000–1.000) & 1.000 (1.000–1.000) \\
        Recall Class 1 & 1.000 (1.000–1.000) & 1.000 (1.000–1.000) & 1.000 (1.000–1.000) & 1.000 (1.000–1.000) \\
        \bottomrule
    \end{tabular}}
    \caption{Summary of results for the two-clouds setup with pretrained Transformer ($90^\circ$ alignment \ref{fig:init_90}). Values are reported as median (first-third quartile).}
    \label{tab:results_two_clouds_pretrained_90}
\end{table}

\begin{table}[h]
    \centering
    \resizebox{\textwidth}{!}{
    \begin{tabular}{lcccc}
        \toprule
        \textbf{Wasserstein distance (between gaussians)} & \textbf{2.000} & \textbf{4.000} & \textbf{6.000} & \textbf{8.000} \\
        \midrule
        Accuracy point-wise & 0.989 (0.975–0.994) & 0.891 (0.845–0.941) & 0.648 (0.576–0.751) & 0.545 (0.465–0.696) \\
        Computational Time & 24.915 (24.776–25.128) & 25.720 (25.016–27.871) & 31.462 (29.966–34.087) & 39.695 (35.811–42.846) \\
        Matching & 0.580 (0.417–0.747) & 0.198 (0.139–0.326) & 0.188 (0.089–0.480) & 0.246 (0.072–0.767) \\
        Wass Distance & 188.653 (180.495–200.400) & 171.845 (159.995–182.669) & 83.392 (67.961–97.934) & 69.315 (60.775–91.427) \\
        Transformer Distance & 189.285 (182.114–200.824) & 176.041 (167.011–186.959) & 91.514 (80.887–103.197) & 81.455 (70.410–97.981) \\
        Transformer Cost & 16503.893 (14810.700–17915.244) & 19359.172 (17605.839–21709.856) & 2999.150 (2026.750–4520.963) & 2681.086 (1780.571–4839.518) \\
        Monge Gap & 0.869 (0.251–1.757) & 4.913 (3.020–6.560) & 8.290 (4.088–11.517) & 8.421 (1.513–13.328) \\
        Efficiency & 0.012 (0.011–0.013) & 0.009 (0.008–0.010) & 0.030 (0.023–0.047) & 0.032 (0.024–0.043) \\
        Optimality & 0.995 (0.990–0.999) & 0.972 (0.961–0.984) & 0.913 (0.863–0.971) & 0.899 (0.824–0.992) \\
        Best Epoch & 199.000 (194.750–199.000) & 195.500 (187.000–199.000) & 9.000 (4.000–26.250) & 9.500 (4.000–27.000) \\
        Accuracy instance-wise & 1.000 (0.978–1.000) & 1.000 (1.000–1.000) & 1.000 (0.927–1.000) & 0.957 (0.710–1.000) \\
        Recall Class 0 & 1.000 (1.000–1.000) & 1.000 (1.000–1.000) & 1.000 (0.955–1.000) & 1.000 (0.898–1.000) \\
        Recall Class 1 & 1.000 (1.000–1.000) & 1.000 (1.000–1.000) & 1.000 (0.957–1.000) & 0.957 (0.728–1.000) \\
        \bottomrule
    \end{tabular}}
    \caption{Summary of results for the two-clouds setup with pretrained Transformer ($180^\circ$ alignment \ref{fig:init_180}). Values are reported as median (first-third quartile).}
    \label{tab:results_two_clouds_pretrained_180}
\end{table}

\begin{table}[h]
    \centering
    \resizebox{\textwidth}{!}{
    \begin{tabular}{lcccc}
        \toprule
        \textbf{mean Wasserstein distance (between gaussians)} & \textbf{2.000} & \textbf{4.000} & \textbf{6.000} & \textbf{8.000} \\
        \midrule
        Accuracy point-wise & 0.668 (0.634–0.724) & 0.644 (0.545–0.772) & 0.608 (0.516–0.671) & 0.565 (0.497–0.617) \\
        Computational Time & 26.844 (26.492–27.302) & 26.316 (25.897–26.932) & 29.181 (28.321–31.429) & 29.471 (27.898–32.265) \\
        Matching & 0.063 (0.023–0.175) & 0.089 (0.038–0.260) & 0.201 (0.071–0.322) & 0.360 (0.133–0.503) \\
        Wass Distance & 131.505 (115.770–149.362) & 131.525 (114.266–162.050) & 132.518 (109.451–151.715) & 118.808 (104.233–150.136) \\
        Transformer Distance & 148.619 (136.774–160.657) & 146.066 (126.464–174.311) & 142.730 (119.085–161.416) & 127.591 (108.820–154.820) \\
        Transformer Cost & 28625.405 (22790.941–33780.430) & 17960.528 (10714.616–25991.450) & 15113.045 (7354.231–18943.695) & 7197.530 (3652.038–19131.785) \\
        Monge Gap & 17.146 (10.004–21.966) & 11.886 (8.317–14.598) & 9.629 (7.253–12.178) & 5.314 (3.313–7.417) \\
        Efficiency & 0.005 (0.004–0.007) & 0.009 (0.006–0.013) & 0.010 (0.008–0.016) & 0.018 (0.008–0.033) \\
        Optimality & 0.882 (0.855–0.937) & 0.924 (0.891–0.942) & 0.929 (0.911–0.950) & 0.959 (0.943–0.974) \\
        Best Epoch & 166.500 (120.750–193.500) & 82.500 (50.250–151.000) & 82.000 (22.500–105.250) & 17.500 (6.000–111.000) \\
        Accuracy instance-wise & 0.956 (0.750–1.000) & 0.956 (0.822–1.000) & 1.000 (0.822–1.000) & 1.000 (0.906–1.000) \\
        Recall Class 0 & 1.000 (1.000–1.000) & 1.000 (1.000–1.000) & 1.000 (1.000–1.000) & 1.000 (1.000–1.000) \\
        Recall Class 1 & 1.000 (1.000–1.000) & 1.000 (0.867–1.000) & 1.000 (1.000–1.000) & 1.000 (1.000–1.000) \\
        Recall Class 2 & 0.933 (0.333–1.000) & 1.000 (0.733–1.000) & 1.000 (0.800–1.000) & 1.000 (1.000–1.000) \\
        \bottomrule
    \end{tabular}}
    \caption{Summary of results for the three-clouds setup with pretrained Transformer ($120^\circ$ rotation \ref{fig:init_three_120}). Values are reported as median (first-third quartile).}
    \label{tab:results_three_clouds_pretrained_120}
\end{table}

\begin{table}[h]
    \centering
    \resizebox{\textwidth}{!}{
    \begin{tabular}{lcccc}
        \toprule
        \textbf{mean Wasserstein distance (between gaussians)} & \textbf{2.000} & \textbf{4.000} & \textbf{6.000} & \textbf{8.000} \\
        \midrule
        Accuracy point-wise & 0.614 (0.500–0.657) & 0.518 (0.410–0.631) & 0.529 (0.439–0.646) & 0.609 (0.534–0.655) \\
        Computational Time & 25.404 (25.339–25.514) & 25.401 (25.353–25.480) & 25.533 (25.410–25.656) & 25.744 (25.572–26.166) \\
        Matching & 0.174 (0.062–0.488) & 0.156 (0.062–0.310) & 0.374 (0.231–0.464) & 0.491 (0.353–0.678) \\
        Wass Distance & 239.469 (189.570–299.564) & 182.299 (146.500–220.827) & 178.097 (149.895–224.350) & 182.922 (162.067–213.857) \\
        Transformer Distance & 249.409 (197.304–301.274) & 191.840 (152.202–227.362) & 187.182 (154.039–228.571) & 188.875 (167.877–215.219) \\
        Transformer Cost & 31177.090 (23173.527–37799.376) & 24216.896 (13336.128–36170.762) & 29643.571 (20902.259–38453.429) & 30367.662 (20085.533–39575.013) \\
        Monge Gap & 8.160 (4.048–13.322) & 8.709 (6.000–11.966) & 4.534 (2.621–6.955) & 3.031 (1.064–5.353) \\
        Efficiency & 0.008 (0.006–0.010) & 0.008 (0.006–0.013) & 0.007 (0.005–0.009) & 0.006 (0.005–0.010) \\
        Optimality & 0.971 (0.932–0.987) & 0.954 (0.934–0.969) & 0.977 (0.962–0.986) & 0.986 (0.970–0.995) \\
        Best Epoch & 186.000 (146.000–198.000) & 94.000 (43.000–173.250) & 152.000 (105.250–189.000) & 166.000 (102.000–191.250) \\
        Accuracy instance-wise & 0.978 (0.844–1.000) & 0.978 (0.556–1.000) & 1.000 (0.622–1.000) & 1.000 (0.722–1.000) \\
        Recall Class 0 & 1.000 (0.933–1.000) & 1.000 (0.917–1.000) & 1.000 (1.000–1.000) & 1.000 (1.000–1.000) \\
        Recall Class 1 & 1.000 (0.933–1.000) & 1.000 (0.783–1.000) & 1.000 (0.717–1.000) & 1.000 (1.000–1.000) \\
        Recall Class 2 & 1.000 (0.933–1.000) & 1.000 (0.117–1.000) & 1.000 (0.133–1.000) & 1.000 (0.933–1.000) \\
        \bottomrule
    \end{tabular}}
    \caption{Summary of results for the three-clouds setup with pretrained Transformer ($180^\circ$ rotation \ref{fig:init_three_180}). Values are reported as median (first-third quartile).}
    \label{tab:results_three_clouds_pretrained_180}
\end{table}

The algorithm seems to be quite sensible to the initialization of the \textit{dummy data}, as can be seen in the drop of \textbf{Accuracy} and \textbf{Matching}.
Interestingly enough this seems to more noticeable when the classification is simpler (bigger separation).
While the pretraining appears to provide a slight improvement in favorable cases, its dependence on the geometry of the initialization makes the method somewhat fragile: it can work well when the dummy cloud is well aligned, but in more complex settings, where little is known about the structure of the data, constructing an effective initialization may prove challenging.

\chapter{An OT-based alternative to Transformer for tabular data}
\label{sec:ot_model}

In light of the findings in Section~\ref{sec:ot_analysis}, I propose an OT model that leverages Optimal Transport in place of Self-Attention to learn an effective remapping of the input data.

The fundamental idea is the following: instead of directly learning a remapping of the training data, we define a Gaussian distribution for each label, ensuring that the distributions are clearly separated from one another. From these distributions---which I will refer to as \textit{dummy distributions}---we sample a set of points whose cardinality corresponds to that of the training data. This collection of sampled points constitutes the \textit{dummy dataset}. We then solve the OT problem between the original (scaled) data and this dummy dataset.

To generalize the resulting transport plan to the entire input space, we employ a simple MLP. This network is trained by taking the \textit{real data} as input and the corresponding matched \textit{dummy data} as target output. Prediction for a new instance is carried out by computing the distance of its remapped representation from each label centroid, followed by the softmax of the negative distances.

\section{Algorithm}
\begin{algorithm}[H]
\small
\label{alg:ot-classification}
\KwIn{$\mathcal{X} \in \mathbb{R}^{T \times P}$ (T time steps, P features), 
       $\mathcal{Y} \in \{0,1,\dots,K-1\}$ (labels)}
\KwOut{Predicted label probabilities for new input $\mathbf{x}_{\text{new}}$}

\BlankLine

\SetKwBlock{StepOne}{\textbf{Step 1: Generate dummy data.}}{}
\StepOne{
Sample the \textit{dummy dataset} $\mathcal{X}_{\text{dummy}}$ by sampling, for each label $k$, for each time step $t$ from a multivariate Gaussian distribution
\[
\mathbf{x}_{k,t} \overset{\text{ind}}{\sim} \mathcal{N}(\mu_k, \Sigma_k),
\]
where $\mu_k \in \mathbb{R}^{P}$  and $\Sigma_k \in \mathbb{R}^{ P \times P}$.
}

\SetKwBlock{StepTwo}{\textbf{Step 2: Compute optimal transport matching.}}{}
\StepTwo{
Solve the OT problem to find 
\[
T : \mathcal{X} \to \mathcal{X}_{\text{dummy}}
\]
minimizing the transport cost between real and dummy data, independently across labels.
}

\SetKwBlock{StepThree}{\textbf{Step 3: Generalize the mapping with an MLP.}}{}
\StepThree{
Train an MLP $f_\theta$ to approximate $T$, i.e.
\[
f_\theta(\mathbf{x}_i) \approx T(\mathbf{x}_i), \quad \forall \mathbf{x}_i \in \mathcal{X}.
\]
}

\SetKwBlock{StepFour}{\textbf{Step 4: Remap new input.}}{}
\StepFour{
For a new instance $\mathbf{x}_{\text{new}} \in \mathbb{R}^{T \times P}$, compute
\[
\mathbf{x}'_{\text{new}} = f_\theta(\mathbf{x}_{\text{new}}).
\]
}

\SetKwBlock{StepFive}{\textbf{Step 5: Distance to label centroids.}}{}
\StepFive{
Let $\mathbf{c}_k$ be the centroid of label $k$ in $\mathcal{X}_{\text{dummy}}$. Compute
\[
d_k = \sum_{t=1}^T \|\mathbf{x}'_{\text{new},t} - \mathbf{c}_k\|_2, \quad k=0,\dots,K-1.
\]
}

\SetKwBlock{StepSix}{\textbf{Step 6: Compute label probabilities.}}{}
\StepSix{
Apply softmax over inverted distances:
\[
p_k = \frac{\exp(-d_k)}{\sum_{j=0}^{K-1} \exp(-d_j)}.
\]

Return $\mathbf{p} = (p_0,\dots,p_{K-1})$.
}
\end{algorithm}

\section{Experiments}

As with the Transformer, a simulation study was conducted to evaluate the performance of the OT model. A total of 100 repetitions were performed.
Training parameters are shown in \ref{tab:params OT}, results are shown in table \ref{tab:results_three_clouds_OT} and \ref{tab:results_two_clouds_OT}.

\begin{table}[h]
    \centering
    \scriptsize
    \begin{tabular}{ll}
        \toprule
        \textbf{Parameter} & \textbf{Value} \\
        \midrule
        MLP epochs & 50 \\
        MLP units & 32, 32 \\
        Dropout & 0.3 \\
        Learning rate & 0.01 \\
        \bottomrule
    \end{tabular}
    \caption{Training parameters of the OT model}
    \label{tab:params OT}
\end{table}

\begin{table}[h]
    \centering
    \resizebox{\textwidth}{!}{
\begin{tabular}{lcccc}
        \toprule
        \textbf{Metric} & \textbf{2.000} & \textbf{4.000} & \textbf{6.000} & \textbf{8.000} \\
        \midrule
        Accuracy point-wise & 0.762 (0.756–0.767) & 0.916 (0.914–0.917) & 0.977 (0.976–0.977) & 0.998 (0.998–0.999) \\
        Computational Time & 4.429 (4.399–4.594) & 4.470 (4.421–4.628) & 4.500 (4.383–4.646) & 4.416 (4.392–4.504) \\
        Best Epoch & 14.500 (6.000–24.000) & 45.000 (40.000–48.000) & 40.000 (31.000–44.000) & 43.000 (39.750–46.000) \\
        Accuracy instance-wise & 1.000 (1.000–1.000) & 1.000 (1.000–1.000) & 1.000 (1.000–1.000) & 1.000 (1.000–1.000) \\
        Recall Class 0 & 1.000 (1.000–1.000) & 1.000 (1.000–1.000) & 1.000 (1.000–1.000) & 1.000 (1.000–1.000) \\
        Recall Class 1 & 1.000 (1.000–1.000) & 1.000 (1.000–1.000) & 1.000 (1.000–1.000) & 1.000 (1.000–1.000) \\
        \bottomrule
    \end{tabular}}
    \caption{Summary of results for two-label classification without pretraining (alternative model, two clouds). Values are reported as median (first–third quartile).}
    \label{tab:results_two_clouds_OT}
\end{table}

\begin{table}[h]
    \centering
    \resizebox{\textwidth}{!}{
    \begin{tabular}{lcccc}
        \toprule
        \textbf{Metric} & \textbf{2.000} & \textbf{4.000} & \textbf{6.000} & \textbf{8.000} \\
        \midrule
        Accuracy point-wise & 0.619 (0.616–0.624) & 0.811 (0.806–0.814) & 0.914 (0.912–0.919) & 0.974 (0.973–0.977) \\
        Computational Time & 4.005 (3.994–4.024) & 4.062 (4.038–4.088) & 4.121 (4.059–4.199) & 4.080 (4.065–4.106) \\
        Best Epoch & 31.500 (22.000–43.000) & 44.000 (39.750–47.000) & 47.000 (45.000–48.000) & 47.000 (45.000–48.000) \\
        Accuracy instance-wise & 1.000 (0.978–1.000) & 1.000 (1.000–1.000) & 1.000 (1.000–1.000) & 1.000 (1.000–1.000) \\
        Recall Class 0 & 1.000 (1.000–1.000) & 1.000 (1.000–1.000) & 1.000 (1.000–1.000) & 1.000 (1.000–1.000) \\
        Recall Class 1 & 1.000 (0.933–1.000) & 1.000 (1.000–1.000) & 1.000 (1.000–1.000) & 1.000 (1.000–1.000) \\
        Recall Class 2 & 1.000 (1.000–1.000) & 1.000 (1.000–1.000) & 1.000 (1.000–1.000) & 1.000 (1.000–1.000) \\
        \bottomrule
    \end{tabular}}
    \caption{Summary of results for three-label classification without pretraining (alternative model, three clouds). Values are reported as median (first–third quartile).}

    \label{tab:results_three_clouds_OT}
\end{table}

This model achieves the same instance-wise performance as the Transformer, but with a much lower computational cost. In practice, both solving the OT problem and training the neural network are more efficient than training the Transformer. The point-wise classification is less accurate, but this should not be a concern in the kind of contexts where Transformers are usually employed (i.e., tasks with long sequences and many tokens per instance, such as NLP, where the goal is to model the sequence as a whole rather than classify individual points).

The main limitations of the algorithm are twofold.  
First, the initialization of the dummy distributions becomes critical as the number of labels increases. The spatial arrangement of these distributions is important because the model relies on the Euclidean distance as the ground metric. If the label set does not naturally carry a meaningful structure (for example, an ordering), the geometry of the dummy distributions could introduce unwanted bias. A natural way to mitigate this would be to place the dummy distributions so that they are equidistant from one another. However, this becomes increasingly difficult as the number of labels grows, since the dimension required to maintain such a configuration may exceed that of the input space.

A possible alternative would be to train a separate MLP for each label and then use the outputs of these networks as scores in the softmax.  

These limitations are the price to pay for the efficiency gains. The OT model may be harder to apply in more complex scenarios, but it clearly shows that there is room for improvement by leveraging OT to learn effective remappings of the data.

With larger datasets, computing OT once on the full training set may become prohibitive; in such cases, it may be preferable to split the data into batches and compute OT between each batch and a dummy set of matching size; this was done in the following application to real data.

\section{Application to real data}

To gain further insight into the practical applicability of the proposed OT model, I consider a real-world dataset: the \textbf{Bangalore EEG Epilepsy Dataset} \footnote{\url{https://archive.ics.uci.edu/dataset/1134/beed:+bangalore+eeg+epilepsy+dataset}}. This dataset contains EEG recordings collected from subjects in Bangalore, India, with the aim of detecting epileptic seizures from non-invasive measurements.  

The dataset has been preprocessed and consists of 8,000 instances, each represented by 16 numerical features extracted from the raw EEG signals. The original labeling scheme includes four classes: healthy subjects (\texttt{y = 0}), generalized seizures (\texttt{y = 1}), focal seizures (\texttt{y = 2}), and seizure-related activities such as eye blinking or staring (\texttt{y = 3}).

I conducted three experiments with increasing label complexity: first, a binary setting where the data were dichotomized into epileptic (\texttt{y = 1}) versus non-epileptic (\texttt{y = 0}); second, a three-class setting using labels \texttt{0–2}; and finally, the full four-class setting.
For each experiment, 20\% of the dataset was reserved as a test set, and the accuracy was computed on this subset.
The pretrained Transformer was trained exclusively on the binary (two-label) setting, as properly initializing the \textit{dummy data} proved challenging for the multi-class configurations.
The results are shown in table \ref{tab:beed_all_labels}.

\begin{table}[h]
\centering
\caption{Comparison of model parameters and performance on the Bangalore EEG dataset for different label settings.}
\resizebox{\textwidth}{!}{%
\begin{tabular}{lccc}
\toprule
 & \textbf{OT model} & \textbf{Transformer} & \textbf{Pretrained Transformer} \\
\midrule
\multicolumn{4}{c}{\textit{Parameters}} \\
Batch size (OT)         & 50  & (-) & (-) \\
Transformer blocks      & (-) & $n_{\text{labels}}-1$ & $n_{\text{labels}}-1$ \\
Number of heads         & (-) & 10  & 10  \\
Head dimension          & (-) & 64  & 64  \\
Feedforward dimension (SA) & (-) & 32 & 32  \\
MLP units               & (32,32) &  (32,32) & (32,32) \\
Dropout (SA)            & 0.3 & 0.1 & 0.1 \\
Dropout (MLP)           & 0.3 & 0.1 & 0.1 \\
Batch size (NN)         & 64  & 64  & 64 \\
Epochs                  & 50  & 100 & 100 \\
Learning rate           & 0.001 & 0.001 & 0.001 \\
\midrule
\multicolumn{4}{c}{\textit{Performance (test set)}} \\
\textbf{2 labels}       & & & \\
Accuracy                & 1.000 & 0.925 & 0.544 \\
Recall (non-epileptic) & 1.000 & 1.000 & 0.619 \\
Recall (epileptic)     & 1.000 & 0.898 & 0.517 \\
Computational time     & 8.62 s & 20.59 s & 20.79 s \\
\midrule
\textbf{3 labels}       & & & \\
Accuracy                & 0.983 & 0.783 & (-) \\
Recall (class 0)       & 1.000 & 1.000 & (-) \\
Recall (class 1)       & 1.000 & 0.975 & (-) \\
Recall (class 2)       & 0.951 & 0.390 &(-) \\
Computational time     & 5.01 s & 29.84 s & (-) \\
\midrule
\textbf{4 labels}       & & & \\
Accuracy                & 0.888 & 0.525 & (-) \\
Recall (class 0)       & 1.000 & 1.000 & (-) \\
Recall (class 1)       & 0.953 & 0.372 & (-) \\
Recall (class 2)       & 0.850 & 0.525 & (-) \\
Recall (class 3)       & 0.714 & 0.143 & (-) \\
Computational time     & 5.72 s & 54.16 s & (-) \\
\bottomrule
\label{tab:beed_all_labels}
\end{tabular}%
}
\end{table}

The OT model achieves the best performance on the Bangalore EEG dataset, with perfect accuracy and recall in the 2-label setting and the lowest computational time. Its success likely benefits from the structured, preprocessed nature of the EEG data, where Euclidean distances remain informative across features.

As the number of labels increases, the performance of the OT model remains consistently high, with only a modest decline in accuracy when moving to three and four classes. By contrast, the Transformer shows a marked drop in performance and becomes increasingly unstable. In particular, while the Transformer always manages to recall class 0, it struggles with the later labels: classes 2 and 3 are often misclassified, with recalls dropping sharply as the label space expands.

Overall, the OT model not only runs faster but also scales better with the number of labels, maintaining reliable performance where the Transformer deteriorates. Much of this advantage likely stems from the structure of the dataset, where each variable has comparable variability and is centered around zero. Such conditions are not always present, but they can often be enforced through standardization, or in higher-dimensional contexts, through the use of embeddings.

These results confirm that the OT model retains strong performance not only in controlled simulations but also in practical biomedical data, where preprocessing and standardized feature structure make Euclidean distances meaningful.

\chapter{Conclusions}

This thesis presents a novel perspective on self-attention by analyzing the remappings it induces on the training data through the lens of Optimal Transport. Self-attention can be interpreted as progressively rearranging inputs into configurations more favorable for prediction. By extracting these remappings at each epoch of gradient descent and comparing them to Optimal Transport solutions, their effectiveness and efficiency were tested.

Traditionally, training neural models has been regarded as an optimization problem in the parameter space. Employing Optimal Transport shifts this view toward the flow of data itself: training can be understood as searching for a transformation of the inputs that best supports the predictive task. This transformation does not necessarily need to be learned; if it can be chosen beforehand, then what remains is to learn the function that links the input data to this final transformation—a much simpler problem to solve.

The analysis shows that the final remapping learned by self-attention is generally very close to optimal, as the point-to-point mapping of Self-Attention nearly coincides with the corresponding Optimal Transport solution. However, the trajectory by which this mapping is reached during training is highly inefficient.

First, I propose training the Attention block and the MLP block separately. The MLP block is trained on dummy data that are easy to classify; once trained, it is frozen, and the self-attention block is then trained on the real data. This procedure mitigates some inefficiencies but remains highly sensitive to the initialization of the dummy data.

Second, I introduced a model that generates dummy data for each label and computes the Optimal Transport plan matching the real data to this dummy set. The mapping is then generalized with a simple MLP, providing an efficient alternative that avoids directly learning the full remapping. This model substantially improves training efficiency while maintaining strong predictive performance and is far less sensitive to the initialization of the dummy data.

Operating in the data space offers significant potential for enhancing the efficiency of Transformer architectures, in those tasks where the geometry of the input space can be exploited Optimal Transport can widely improve the efficiency of the Transformer.
Questions remain on how many tasks fall in this category, but because of the versatility of discrete Optimal Transport there is reason to believe that this method could be widely applicable.

The drawback of this approach is that it often requires a task-specific design of the algorithm. In contrast, the standard Transformer requires minimal modifications across a wide range of tasks, highlighting a tradeoff between efficiency and transferability.

Future improvements could focus on settings with a large number of labels, where managing the geometry of the dummy distributions becomes more challenging. The approach could also be extended to other tasks, particularly regression.

In summary, this work demonstrates that incorporating Optimal Transport into Transformer training can yield substantial efficiency gains without compromising predictive performance. This generally comes at the expense of flexibility and transferability.

\clearpage

\appendix

\chapter{Push-Forward Operator}
\label{appendix:pushforward}

Given a measurable map $T:\mathcal{X} \to \mathcal{Y}$ and a measure $\mu \in \mathcal{M}_+(\mathcal{X})$, the \emph{push-forward measure} $T_{\#}\mu \in \mathcal{M}_+(\mathcal{Y})$ is defined by
\[
    T_{\#}\mu(B) = \mu(T^{-1}(B)) \qquad \text{for all measurable } B \subseteq \mathcal{Y}.
\]
Equivalently, for any $h \in C(\mathcal{Y})$,
\[
    \int_{\mathcal{Y}} h(y) \, d(T_{\#}\mu)(y) 
    = \int_{\mathcal{X}} h(T(x)) \, d\mu(x).
\]
This operator formalizes the idea of transporting the entire distribution $\mu$ through the map $T$. In the discrete case, if $\mu = \sum_i a_i \delta_{x_i}$, then
\[
    T_{\#}\mu = \sum_i a_i \delta_{T(x_i)}.
\]

\chapter{Computational Environment}
\label{app:computational_environment}

All simulations and experiments reported in this thesis were conducted on a MacBook Pro (Intel x86\_64) running macOS Big Sur. Experiments were executed in R 4.2.2 (2022-10-31) using CPU only (no GPU acceleration). The following session information provides a complete snapshot of the software environment and attached packages, ensuring reproducibility.

\begin{verbatim}
R version 4.2.2 (2022-10-31)
Platform: x86_64-apple-darwin17.0 (64-bit)
Running under: macOS Big Sur 10.16

Matrix products: default
BLAS:   /Library/Frameworks/R.framework/Versions/4.2/Resources/lib/libRblas.0.dylib
LAPACK: /Library/Frameworks/R.framework/Versions/4.2/Resources/lib/libRlapack.dylib

locale:
[1] C/UTF-8/C/C/C/C

attached base packages:
[1] stats     graphics  grDevices utils     datasets  methods   base     

other attached packages:
[1] transport_0.14-6  expm_0.999-8      Matrix_1.5-4      mvtnorm_1.2-4    
[5] dplyr_1.1.3       readr_2.1.4       keras3_1.4.0      tensorflow_2.16.0

loaded via a namespace (and not attached):
 [1] Rcpp_1.0.12       compiler_4.2.2    pillar_1.9.0      base64enc_0.1-3  
 [5] tools_4.2.2       zeallot_0.1.0     jsonlite_1.8.8    lifecycle_1.0.4  
 [9] tibble_3.2.1      lattice_0.21-8    pkgconfig_2.0.3   png_0.1-8        
[13] rlang_1.1.2       cli_3.6.2         dotty_0.1.0       generics_0.1.3   
[17] vctrs_0.6.5       hms_1.1.3         rprojroot_2.0.4   grid_4.2.2       
[21] tidyselect_1.2.1  reticulate_1.43.0 glue_1.7.0        data.table_1.15.4
[25] here_1.0.1        R6_2.5.1          fansi_1.0.4       tzdb_0.4.0       
[29] magrittr_2.0.3    whisker_0.4.1     codetools_0.2-19  tfruns_1.5.3     
[33] utf8_1.2.3
\end{verbatim}

\noindent This environment ensures that all results are reproducible and provides full transparency about the software versions and dependencies used in the analyses.

\cleardoublepage

\fancyhf{} 
\fancyhead[LE,RO]{\thepage \sffamily}
\fancyhead[LO,RE]{\textit{Bibliografia}}

\bibliography{biblio}

@article{vaswani2017attention,
  title={Attention is all you need},
  author={Vaswani, Ashish and Shazeer, Noam and Parmar, Niki and Uszkoreit, Jakob and Jones, Llion and Gomez, Aidan N and Kaiser, {\L}ukasz and Polosukhin, Illia},
  journal={arXiv preprint arXiv:1706.03762},
  year={2017},
  url={https://arxiv.org/abs/1706.03762}
}

@article{monge_gap,
  title={The Monge Gap: A Regularizer to Learn All Transport Maps},
  author={Uscidda, Th{\'e}o and Cuturi, Marco},
  journal={Proceedings of the 40th International Conference on Machine Learning (ICML)},
  year={2023},
  volume={202},
  pages={12345--12356},
  doi={10.48550/arXiv.2302.04953},
  url={https://arxiv.org/abs/2302.04953}
}

@article{sander2022sinkformers,
  title={Sinkformers: Transformers with Doubly Stochastic Attention},
  author={Sander, Michael E. and Ablin, Pierre and Blondel, Mathieu and Peyr{\'e}, Gabriel},
  journal={arXiv preprint arXiv:2110.11773},
  year={2022},
  url={https://arxiv.org/abs/2110.11773}
}

@Article{drones7050287,
AUTHOR = {Jamil, Sonain and Jalil Piran, Md. and Kwon, Oh-Jin},
TITLE = {A Comprehensive Survey of Transformers for Computer Vision},
JOURNAL = {Drones},
VOLUME = {7},
YEAR = {2023},
NUMBER = {5},
ARTICLE-NUMBER = {287},
URL = {https://www.mdpi.com/2504-446X/7/5/287},
ISSN = {2504-446X},
DOI = {10.3390/drones7050287}
}

@misc{transformersspeechprocessingsurvey,
      title={Transformers in Speech Processing: A Survey}, 
      author={Siddique Latif and Aun Zaidi and Heriberto Cuayahuitl and Fahad Shamshad and Moazzam Shoukat and Muhammad Usama and Junaid Qadir},
      year={2025},
      eprint={2303.11607},
      archivePrefix={arXiv},
      primaryClass={cs.CL},
      url={https://arxiv.org/abs/2303.11607}, 
}

@ARTICLE{multimodal,
  author={Xu, Peng and Zhu, Xiatian and Clifton, David A.},
  journal={IEEE Transactions on Pattern Analysis and Machine Intelligence}, 
  title={Multimodal Learning With Transformers: A Survey}, 
  year={2023},
  volume={45},
  number={10},
  pages={12113-12132},
  doi={10.1109/TPAMI.2023.3275156}}

@inproceedings{pruning,
 author = {Kwon, Woosuk and Kim, Sehoon and Mahoney, Michael W and Hassoun, Joseph and Keutzer, Kurt and Gholami, Amir},
 booktitle = {Advances in Neural Information Processing Systems},
 editor = {S. Koyejo and S. Mohamed and A. Agarwal and D. Belgrave and K. Cho and A. Oh},
 pages = {24101--24116},
 publisher = {Curran Associates, Inc.},
 title = {A Fast Post-Training Pruning Framework for Transformers},
 url = {https://proceedings.neurips.cc/paper_files/paper/2022/file/987bed997ab668f91c822a09bce3ea12-Paper-Conference.pdf},
 volume = {35},
 year = {2022}
}

@inproceedings{distilleria,
  title={Training data-efficient image transformers \& distillation through attention},
  author={Touvron, Hugo and Cord, Matthieu and Douze, Matthijs and Massa, Francisco and Sablayrolles, Alexandre and J{\'e}gou, Herv{\'e}},
  booktitle={International conference on machine learning},
  pages={10347--10357},
  year={2021},
  organization={PMLR}
}

@article{compression,
  title={A survey on transformer compression},
  author={Tang, Yehui and Wang, Yunhe and Guo, Jianyuan and Tu, Zhijun and Han, Kai and Hu, Hailin and Tao, Dacheng},
  journal={arXiv preprint arXiv:2402.05964},
  year={2024}
}

@article{peyre2019computational,
  title={Computational optimal transport: With applications to data science},
  author={Peyr{\'e}, Gabriel and Cuturi, Marco and others},
  journal={Foundations and Trends{\textregistered} in Machine Learning},
  volume={11},
  number={5-6},
  pages={355--607},
  year={2019},
  publisher={Now Publishers, Inc.}
}

@inproceedings{arjovsky2017wasserstein,
  title={Wasserstein generative adversarial networks},
  author={Arjovsky, Martin and Chintala, Soumith and Bottou, L{\'e}on},
  booktitle={International conference on machine learning},
  pages={214--223},
  year={2017},
  organization={PMLR}
}

@article{global,
  title={Global convergence in training large-scale transformers},
  author={Gao, Cheng and Cao, Yuan and Li, Zihao and He, Yihan and Wang, Mengdi and Liu, Han and Klusowski, Jason and Fan, Jianqing},
  journal={Advances in Neural Information Processing Systems},
  volume={37},
  pages={29213--29284},
  year={2024}
}

@article{OTambrogio,
  title={Optimal transport for applied mathematicians},
  author={Santambrogio, Filippo},
  year={2015},
  publisher={Springer}
}

@article{santambrogio2017euclidean,
  title={$\{$Euclidean, metric, and Wasserstein$\}$ gradient flows: an overview},
  author={Santambrogio, Filippo},
  journal={Bulletin of Mathematical Sciences},
  volume={7},
  number={1},
  pages={87--154},
  year={2017},
  publisher={Springer}
}

@article{lambert2022variational,
  title={Variational inference via Wasserstein gradient flows},
  author={Lambert, Marc and Chewi, Sinho and Bach, Francis and Bonnabel, Silv{\`e}re and Rigollet, Philippe},
  journal={Advances in Neural Information Processing Systems},
  volume={35},
  pages={14434--14447},
  year={2022}
}

@book{aggarwal2018neural,
  title={Neural networks and deep learning},
  author={Aggarwal, Charu C and others},
  volume={10},
  number={978},
  year={2018},
  publisher={Springer}
}

@article{badaro2023transformers,
  title={Transformers for tabular data representation: A survey of models and applications},
  author={Badaro, Gilbert and Saeed, Mohammed and Papotti, Paolo},
  journal={Transactions of the Association for Computational Linguistics},
  volume={11},
  pages={227--249},
  year={2023},
  publisher={MIT Press One Broadway, 12th Floor, Cambridge, Massachusetts 02142, USA~…}
}

@article{peyre2025optimal,
  title={Optimal Transport for Machine Learners},
  author={Peyr{\'e}, Gabriel},
  journal={arXiv preprint arXiv:2505.06589},
  year={2025}
}

\end{document}